\documentclass[letterpaper]{article} 
\usepackage{aaai2027}  
\usepackage[hyphens]{url}  
\usepackage{graphicx} 
\usepackage{natbib}  
\usepackage{caption} 
\usepackage{algorithm}
\usepackage{algorithmic}
\usepackage{newfloat}
\usepackage{listings}
\DeclareCaptionStyle{ruled}{labelfont=normalfont,labelsep=colon,strut=off} 
\floatstyle{ruled}
\newfloat{listing}{tb}{lst}{}
\floatname{listing}{Listing}

\usepackage{booktabs}

\title{PATS: Policy-Aware Training Scaffolding for Agentic Reinforcement Learning}
\usepackage{xspace}
\usepackage{amsmath}
\usepackage{amssymb}
\newcommand{\method}{\textsc{Pats}\xspace}

\author {
    Yipeng Shi\textsuperscript{\rm 1,2}\equalcontrib,
    Zhipeng Ma\textsuperscript{\rm 2}\equalcontrib,
    Yue Wang\textsuperscript{\rm 2},
    Qitai Tan\textsuperscript{\rm 2},
    Yang Li\textsuperscript{\rm 2},
    Peng Chen\textsuperscript{\rm 2},
    Zhengzhou Zhu\textsuperscript{\rm 1}\corresponding
}
\affiliations {
    \textsuperscript{\rm 1}School of Software \& Microelectronics, Peking University\\
    \textsuperscript{\rm 2}Tencent\\
    ypshi25@stu.pku.edu.cn, 
    \{allenzpma, appleywang, qitaitan, thomasyngli, pengchen\}@tencent.com, 
    zhuzz@ss.pku.edu.cn
    
}

\begin{document}

\maketitle  

\begin{abstract}
In long-horizon LLM agent reinforcement learning, weak policies often repeat similar failures, producing uninformative rollout trajectories and limiting effective policy optimization. Existing skill-centric methods improve exploration by optimizing, filtering, or internalizing reusable skills. However, they remain centered on the skills themselves rather than being designed as adaptive training-time support for the evolving policy. To address this, we propose a policy-centric training paradigm that reframes skills as a dynamic training scaffold. Our framework, \method, converts rollout groups from the latest policy into evidence cards and uses task-specific evaluation to adjust the context used in subsequent rollouts. Concrete guidance helps weak policies to complete challenging tasks. As policy improves, redundant context is revised or removed to reduce reliance on explicit guidance while preserving useful rollout variation. The policy is optimized with environmental rewards using standard RLVR, and the training scaffold is discarded at deployment. Across ALFWorld, WebShop, and seven search-augmented QA benchmarks, \method achieves performance competitive with SOTA baselines while using 25\%--50\% fewer tokens.

\end{abstract}


\begin{links}
    \link{Code}{https://github.com/shi-yipeng/PATS}
\end{links}
\section{Introduction}
Large language model (LLM) agents \citep{react2022,alfworld2020,webshop2022} have demonstrated standard-setting progress in tackling complex, multi-step decision-making problems. Reinforcement learning (RL) can improve such agents, but its efficacy is bounded by policy where rewards are obtainable. In long-horizon agentic tasks, weak policies often repeatedly suffer from identical failures. Consequently, group-relative methods such as GRPO fail to derive sufficient supervisory signals from the rollout groups to drive policy optimization \citep{deepseekmath2024}.

External skills can help a weak policy reach otherwise inaccessible capabilities \citep{reflexion2023,expel2023,voyager2023,skillrl2026}.However, skills that are effective during inference are not always beneficial for training. While inference prioritizes hints that maximize immediate success, training requires differentiated trajectories that reveal effective policy updates. Specifically, at different stages of training, insufficient hints can lead to systemic failures across certain tasks, whereas overly comprehensive hints risk suppressing policy generalization. The key question is not which skills maximize the model's instant performance, but rather: \emph{what kind of skills does the current policy need to generate more informative rollout groups for policy optimization?}

\begin{figure}[h]
\centering
\vspace{-1.0em}
\includegraphics[width=0.95\columnwidth]{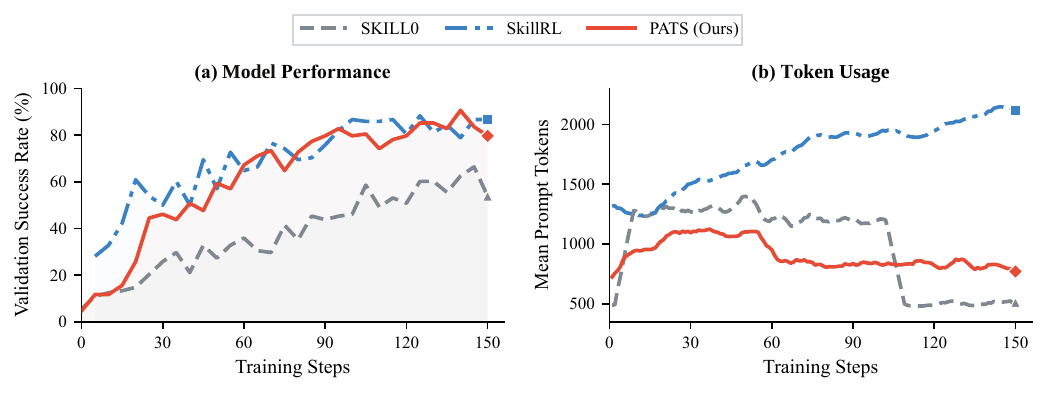}
\caption{Training dynamics on ALFWorld under the shared 150-step RL budget. }
\label{fig:alfworld-training-dynamics}
\vspace{-1.0em} 
\end{figure}

\begin{figure*}[th]
\centering
\includegraphics[width=0.49\textwidth]{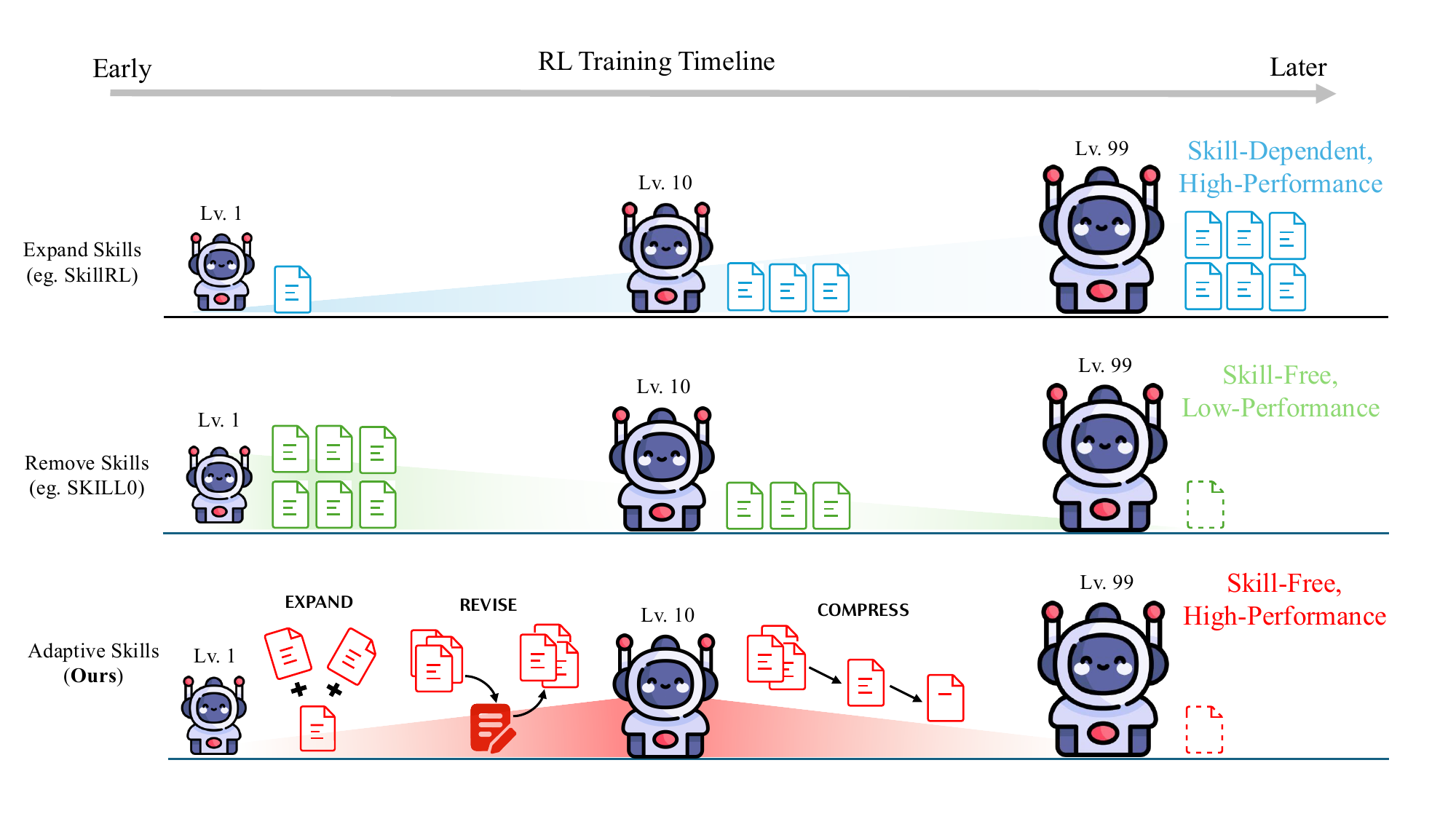}\hfill
\includegraphics[width=0.49\textwidth]{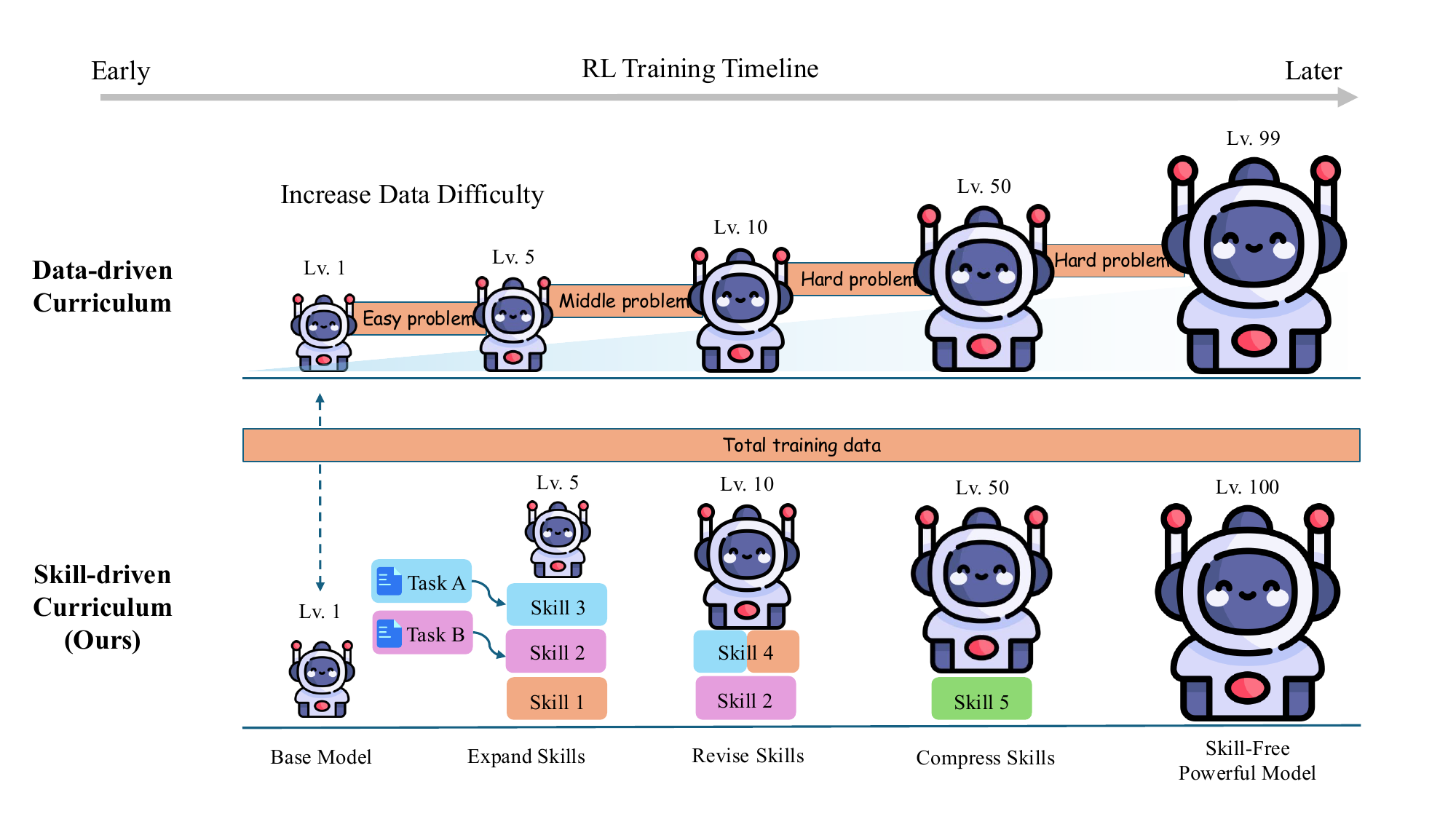}
\caption{Comparison of different training methods. Left: Skill-expansion methods retain a growing set of experiences, whereas skill-removal methods reduce the number of skills to internalize task competence. \method dynamically adjusts skills through expansion, semantic revision, compression, and bounded pruning. Right: Data-driven curricula alter task difficulty, whereas \method leaves the dataset uncontrolled and only modulates temporary skills using training scaffold.}
\label{fig:intro-overview}
\end{figure*}

To address this challenge, we introduce \method, a framework that reframes skills not as static prompt extensions, but as a \emph{disposable, policy-conditioned training scaffold}.  \method initializes with an empty scaffold and convert these rollout group into evidence cards. By contrasting successful and unsuccessful trajectories, these cards capture the explicit long-term knowledge acquired during rollouts to continuously update the scaffolded skills. Following prior work \citep{divagrpo2026}, maximizing training efficiency requires maintaining a rollout success rate around 0.5—the regime of maximum exploration contrast. Therefore, the scaffold dynamically applies differentiated skill operations tailored to varying rollout performance. To smooth out sample-level difficulty noise, we aggregate tasks by category and employ category-level success rates as control signals to derive control primitives. Driven by these primitives, a frozen refiner model expands, revises, or compresses skills within the scaffold to match the policy's evolving learning needs. The updated skills are then inject to guide the subsequent rollout phase.

Figure~\ref{fig:alfworld-training-dynamics} shows the intended training scaffolding pattern: skills expand when the weak policy needs concrete guidance and contract as performance improves. This confirms that the performance gains stem from adaptive training support rather than reliance on well-crafted or continuously evolving skills.

To contextualize our contribution, we evaluate \method against existing paradigms (Figure~\ref{fig:intro-overview}),including monotonic skill expansion, monotonic skill removal, and traditional data-driven curriculum learning. Specifically, while monotonic skill expansion accumulates prompts to yield high performance at the cost of persistent skill dependence, monotonic skill removal fails to achieve competitive performance despite becoming skill-free. In contrast, our adaptive training scaffold dynamically expands, revises, and compresses skills throughout RL training, ultimately enabling a \textbf{skill-free, high-performance} model. Furthermore, unlike data-driven curricula that manually schedule task difficulties, our skill-driven paradigm leaves the underlying training dataset uncontrolled, modulating dynamic scaffolding alone to achieve effective curriculum learning.

In summary, our main contributions include:
\begin{itemize}
    \item We propose a training paradigm named \emph{Training Scaffolding}, which models historical experience as a temporary skill during rollouts. As the policy improves, skills dynamically adapt to guide policy updates, ultimately enabling skill-free deployment.
    
    \item We design a policy-aware scaffold adaptation framework. The framework converts rollout groups into task-level evidence cards, applies primitive-driven refinements to update current skills using these cards, and merges the updated slices through atomic commits. Crucially, this component executes in parallel with policy optimization with no extra time cost.
    
    \item Extensive experiments on ALFWorld, WebShop, and seven search-augmented QA benchmarks show that \method delivers performance competitive with SOTA baseline (varying within $\pm 3\%$), while reducing token consumption by 25\%--50\%.
\end{itemize}

\section{Related Work}

\paragraph{Agentic reinforcement learning.}
Agentic RL methods often enhance learning by refining policy objectives and credit assignment for multi-turn trajectories \citep{ppo2017,rloo2024,deepseekmath2024,gigpo2025,ragen2025}.Curricula instead change which problems the policy encounters, using difficulty or group reward variance to avoid uniformly easy or impossible samples \citep{curriculum2009,vcrl2025,divagrpo2026}. Other methods shape rollouts directly: ActGuide-RL supplies plan-like action guidance as an adaptive fallback, whereas Rubric-Scaffolded RL uses rubrics to guide both generation and reward construction \citep{actguiderl2026,ruscarl2025}. \method operates on a complementary control surface. It keeps the task distribution, environment reward, and GRPO objective fixed, and adapts only the experience-derived context supplied to subsequent rollout groups.

\vspace{-0.5em}
\paragraph{Skill artifacts, adaptation, and internalization.}
Agent memories and skills turn past interaction into reusable reflections, procedures, or executable behaviors \citep{reflexion2023,expel2023,voyager2023}. SkillRL evolves a retrieved SkillBank during RL, while SkillOS and Skill1 explicitly optimize skill curation, selection, use, or distillation \citep{skillrl2026,skillos2026,skillone2026}. A second line studies when external skills should be retained or absorbed: SKILL0 schedules their withdrawal, SLIM estimates leave-one-skill-out contributions, and SkillC and SIRI compare supported and unsupported behavior for credit assignment or distillation \citep{skillzero2026,slim2026,skillc2026,siri2026}. SEED converts hindsight skills into token-level on-policy distillation, while UCOB treats skill-conditioned and skill-free prompts as paired policy views and uses their local return difference for distillation and memory updates \citep{seed2026,ucob2026}. While existing methods treat skills as optimization targets, \method views them as a disposable training scaffold. By leveraging group evidence, the framework dynamically adjusts support for future rollouts without requiring per-skill validation, auxiliary distillation, or test-time retrieval.

\section{Methodology}
\label{sec:pats}

\begin{figure*}[t]
\centering
\includegraphics[width=0.85\textwidth]{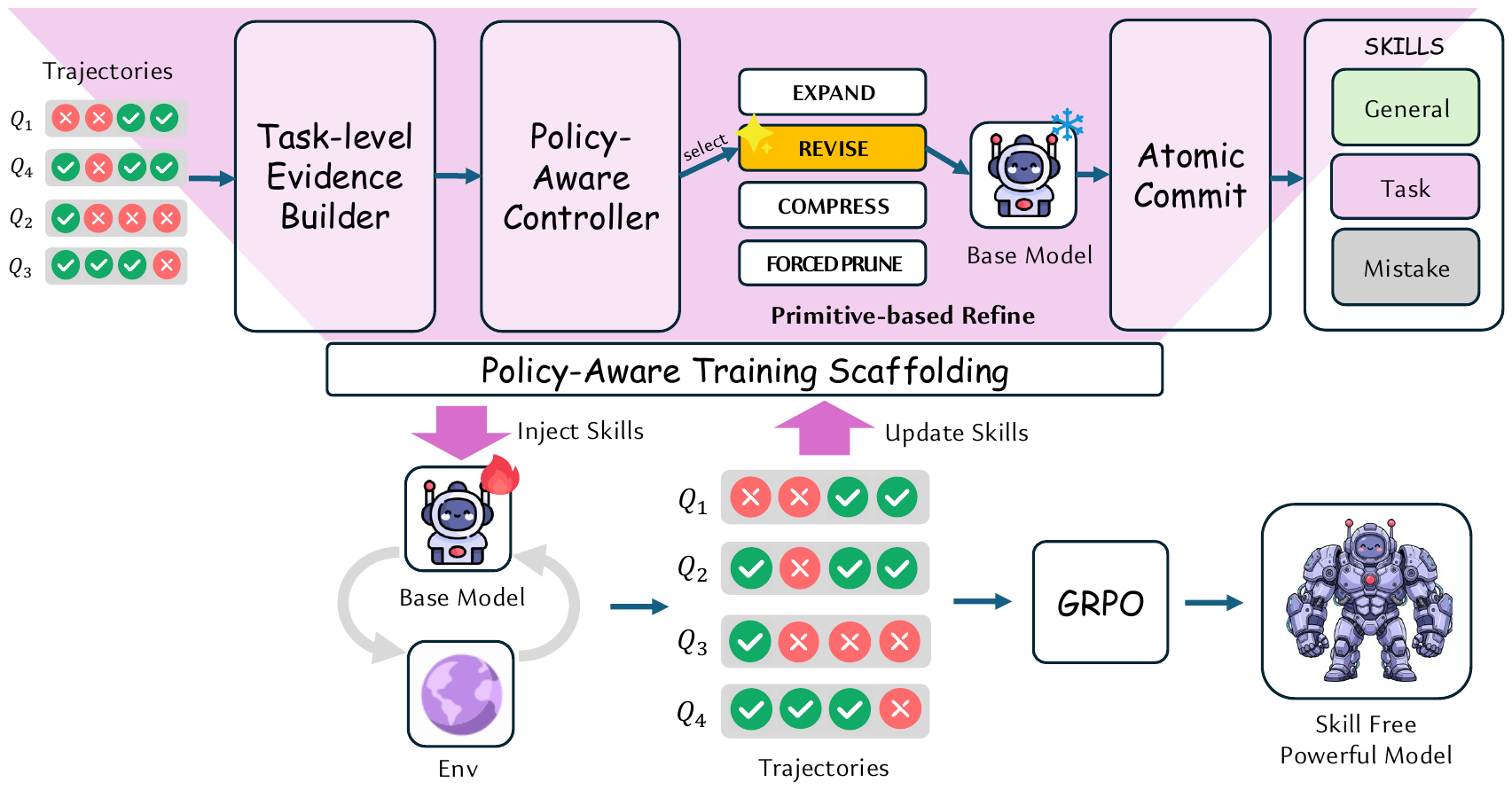}
\caption{Overview of \method. Skills are injected into rollout requests as context, and the resulting trajectories are used to update the policy via GRPO. Policy updates and skill updates are executed in parallel. The updated skills take effect in the subsequent iteration and are discarded during deployment.}
\label{fig:method}
\end{figure*}

\subsection{Overview}

We study interactive tasks drawn from distribution $\mathcal{D}$. Each task $q$ belongs to a task type $\tau(q)$ in the set $\mathcal{T}$. An agent may receive temporary textual support during training but must act without that support at deployment. Given observation $o_t$, bounded interaction history $h_t$, and action $a_t$, a complete trajectory is
\begin{equation}
\xi=(q,o_1,a_1,\ldots,o_T,a_T),
\end{equation}
where $T$ is determined by task completion, failure, or the interaction limit. The environment returns an outcome reward $r(\xi)$ after the trajectory ends. Although \method uses a structured scaffold $\mathcal{B}$ during training, deployment removes this scaffold:
\begin{equation}
J_{\mathrm{deploy}}(\theta)
=
\mathbb{E}_{q\sim\mathcal{D}}
\mathbb{E}_{\xi\sim\pi_\theta(\cdot\mid q,\varnothing)}
\left[r(\xi)\right].
\label{eq:deploy-objective}
\end{equation}
Thus, the scaffold is not optimized as a persistent memory artifact. It is a training-time support variable whose content, specificity, and cost should change as the policy becomes more competent.

\method evolves the policy and scaffold on separate but coupled paths. At RL iteration $k$, all rollouts read a frozen scaffold snapshot $\widetilde{\mathcal{B}}_k$. The resulting trajectory batch $\Xi_k$ updates the policy through standard reinforcement learning and also produces evidence for future scaffold edits:
\begin{align}
\theta_{k+1}
&=\operatorname{RLUpdate}
\left(\theta_k;\,\Xi_k,\widetilde{\mathcal{B}}_k\right), \\
\mathcal{B}_{k+1}
&=\operatorname{ScaffoldUpdate}
\left(\mathcal{B}_k;\,\operatorname{Evidence}(\Xi_k),
\{u_{\tau,k}\}_{\tau\in\mathcal{T}_k}\right),
\label{eq:two-path-update}
\end{align}
where $\mathcal{T}_k\subseteq\mathcal{T}$ contains the task types sampled at iteration $k$, and $u_{\tau,k}$ is the controller state defined below. The second path is a discrete training-state update rather than a differentiable optimization path. Because every rollout in iteration $k$ reads the frozen snapshot $\widetilde{\mathcal{B}}_k$, edits derived from $\Xi_k$ cannot affect the batch that produced them. Validated edits become visible only in the next snapshot.

Algorithm~\ref{alg:pats} summarizes the procedure. Each iteration renders a task-type scaffold view, samples parallel scaffold-conditioned trajectories, updates the policy with environmental rewards, aggregates rollout evidence by task type, selects an edit mode from competence and scaffold pressure, and commits only validated edits.

\begin{algorithm}[t]
\caption{\method: Policy-Aware Training Scaffolding}
\label{alg:pats}
\begin{algorithmic}[1]
\REQUIRE $\pi_{\mathrm{base}}$, $\mathcal{D}_{\mathrm{SFT}}$, $\mathcal{D}$, initial scaffold $\mathcal{B}_0$, refiner $R$
\ENSURE Deployment policy $\pi_{\theta^\star}(\cdot\mid q,\varnothing)$ without the training scaffold
\STATE $\theta\leftarrow\operatorname{SFT}(\pi_{\mathrm{base}},\mathcal{D}_{\mathrm{SFT}})$
\STATE $\mathcal{B}\leftarrow\mathcal{B}_0$
\FOR{training iteration $k=0,\ldots,K-1$}
  \STATE $\widetilde{\mathcal{B}}_k\leftarrow\operatorname{Snapshot}(\mathcal{B})$
  \STATE $\Xi_k\leftarrow\varnothing,\quad\mathcal{E}_k\leftarrow\varnothing$
  \FORALL{sampled task $q$ with task type $\tau$}
    \STATE $z_{\tau,k}\leftarrow\operatorname{RetrieveRender}(\tau,\widetilde{\mathcal{B}}_k)$
    \STATE $G_{q,k}\leftarrow\{\xi_i\}_{i=1}^{n}\sim\pi_\theta(\cdot\mid q,z_{\tau,k})$
    \STATE $\Xi_k\leftarrow\Xi_k\cup G_{q,k}$
    \STATE $\mathcal{E}_k[\tau]\leftarrow\mathcal{E}_k[\tau]\cup\operatorname{BuildEvidence}(G_{q,k})$
  \ENDFOR
  \STATE $\theta\leftarrow\operatorname{GroupRelativeUpdate}(\theta;\Xi_k)$
  \FORALL{task type $\tau$ represented in $\mathcal{E}_k$}
    \STATE $c_{\tau,k}\leftarrow\operatorname{Aggregate}(\mathcal{E}_k[\tau])$
    \STATE $u_{\tau,k}\leftarrow\operatorname{UpdateState}(c_{\tau,k},\mathcal{B})$
    \STATE $\mu_{\tau,k}\leftarrow\operatorname{SelectMode}(u_{\tau,k})$
    \STATE $\mathcal{O}_{\tau,k}\leftarrow R(c_{\tau,k},\mathcal{B}_{\tau},u_{\tau,k},\mu_{\tau,k})$
    \STATE $\widehat{\mathcal{O}}_{\tau,k}\leftarrow\operatorname{Validate}(\mathcal{B},\mathcal{O}_{\tau,k},\mu_{\tau,k})$
    \STATE $\mathcal{B}\leftarrow\operatorname{CommitApply}(\mathcal{B},\widehat{\mathcal{O}}_{\tau,k})$
  \ENDFOR
\ENDFOR
\STATE $\theta^\star\leftarrow\theta$
\RETURN $\pi_{\theta^\star}(\cdot\mid q,\varnothing)$
\end{algorithmic}
\end{algorithm}

\subsection{Scaffold-Conditioned Policy Optimization}

\method stores experience as structured textual entries rather than full trajectories. Following SkillBank-inspired organization \citep{anthropicagentskills2025,skillrl2026}, the scaffold contains three semantic layers:
\begin{equation}
\mathcal{B}_k
=
\mathcal{B}_k^{\mathrm{gen}}
\cup
\left(\bigcup_{\tau\in\mathcal{T}}
\mathcal{B}_{k,\tau}^{\mathrm{task}}\right)
\cup
\mathcal{B}_k^{\mathrm{mistake}}.
\label{eq:scaffold-layers}
\end{equation}
General entries encode principles reusable across task types, task entries encode procedures for a specific task type $\tau$, and mistake entries encode recurring failures with corrective actions. Each entry pairs an actionable rule with a natural-language applicability condition. The exact fields, prompt template, and validation schema are given in Appendix~\ref{app:stage1-interface}.

For task type $\tau$, a deterministic renderer selects a bounded view from the frozen snapshot:
\begin{equation}
z_{\tau,k}
=
\operatorname{Render}\left(
\operatorname{Slice}(
\mathcal{B}_k^{\mathrm{gen}}
\cup
\mathcal{B}_{k,\tau}^{\mathrm{task}}
\cup
\mathcal{B}_k^{\mathrm{mistake}})
\right).
\label{eq:rendered-scaffold}
\end{equation}
Retrieval depends on the task type and scaffold snapshot. The policy decides whether an entry applies from the task, history, observation, and the entry's semantic condition:
\begin{equation}
a_t\sim
\pi_{\theta_k}
\left(\cdot\mid q,h_t,o_t,z_{\tau,k}\right).
\label{eq:scaffolded-policy}
\end{equation}
In unsupported evaluation, the entire experience block is removed while the original task renderer, history, observation, action format, and decoding settings are retained. This makes the intended deployment condition explicit: the final policy must use behavior learned during training, not a retrieved skill library.

\subsection{Policy-Aware Scaffold Adaptation}
\label{sec:scaffold-adaptation}

\method adapts the scaffold through three components: group-level evidence summarizes the latest success--failure contrast, a competence controller selects the support level, and a constrained refiner edits the scaffold without changing the current on-policy batch.

\paragraph{Group-level evidence aggregation.}
Updating the scaffold directly from an individual trajectory can overreact to policy sampling noise, environment branching, or an idiosyncratic failure. The basic evidence unit of \method is therefore a GRPO group of parallel attempts on the same concrete task. For task $q$ of type $\tau(q)$, let
\begin{equation}
G_{q,k}=\{\xi_{q,i}\}_{i=1}^{n},
\qquad
s_{q,k}=\frac{1}{n}\sum_{i=1}^{n}\mathbf{1}\{r(\xi_{q,i})=1\},
\label{eq:rollout-group}
\end{equation}
where all $n$ trajectories share $q$ and the frozen scaffold view for $\tau(q)$. We partition them into successful and failed subsets $G_{q,k}^{+}$ and $G_{q,k}^{-}$. A deterministic evidence builder extracts the group success rate, recurring environment feedback in failed trajectories, and action patterns that occur more often in successful trajectories. To keep updates auditable without exposing free-form hidden reasoning, it also selects representative traces containing only observations, executable actions, and environment feedback.

Formally, let $\Phi(G)$ denote deterministic aggregate features and $\operatorname{Rep}(G)$ denote a representative-trace selector. The evidence card is
\begin{equation}
\begin{aligned}
c_{q,k}=\operatorname{Card}\big(&
q,\tau(q),s_{q,k},\Phi(G_{q,k}^{+}),\Phi(G_{q,k}^{-}),\\
&\operatorname{Rep}(G_{q,k}^{+}),\operatorname{Rep}(G_{q,k}^{-})\big).
\end{aligned}
\label{eq:evidence-card}
\end{equation}
Cards are then aggregated by task type:
\begin{equation}
c_{\tau,k}
=
\operatorname{Aggregate}\!\left(
\{c_{q,k}:\tau(q)=\tau\}
\right).
\label{eq:task-type-card}
\end{equation}
Thus, each card first preserves the success--failure contrast within a valid GRPO group, while the task-type aggregate supplies broader evidence to one scaffold update. Evidence cards do not decide whether to add, revise, or delete support; rewards and GRPO advantages remain defined only by the environment.

\paragraph{Primitive-based Skill Refine.}
The amount of useful support changes as the policy learns. Low-competence policies need concrete guidance that makes successful trajectories reachable. At intermediate competence, support should track the policy's residual failures rather than simply continue to grow. High-competence policies should lose redundant support before it becomes a crutch.

Let $\mathcal{Q}_{\tau,k}$ be the sampled tasks of type $\tau$ at iteration $k$. Their mean success rate is
\begin{equation}
s_{\tau,k}
=
\frac{1}{|\mathcal{Q}_{\tau,k}|}
\sum_{q\in\mathcal{Q}_{\tau,k}}s_{q,k}.
\label{eq:task-type-success}
\end{equation}
For each task type, \method tracks an exponential moving average of this quantity:
\begin{equation}
\bar{s}_{\tau,k}
=
\alpha s_{\tau,k}
+(1-\alpha)\bar{s}_{\tau,k-1}.
\label{eq:success-ema}
\end{equation}
It also tracks global scaffold pressure as the tighter of entry-count and rendered-token constraints:
\begin{equation}
p_k
=
\max\left(
\frac{|\mathcal{B}_k|}{N_{\max}},
\frac{\operatorname{Tok}(\mathcal{B}_k)}{L_{\max}}
\right).
\label{eq:scaffold-pressure}
\end{equation}
Here, $|\mathcal{B}_k|$ is the number of scaffold entries, $\operatorname{Tok}(\mathcal{B}_k)$ is their serialized token count, and $N_{\max}$ and $L_{\max}$ are the corresponding entry and token budgets. Thus, $p_k$ is the larger normalized budget utilization and reaches one when either capacity limit is met.
The controller maps the state $u_{\tau,k}=(\bar{s}_{\tau,k},p_k)$ to one of four modes:
\begin{equation}
\mu_{\tau,k}=
\begin{cases}
\mathrm{FORCED\_PRUNE}, & p_k\ge 1,\\
\mathrm{COMPRESS}, & \bar{s}_{\tau,k}\ge\rho_c,\\
\mathrm{REVISE}, & \rho_r\le\bar{s}_{\tau,k}<\rho_c,\\
\mathrm{EXPAND}, & \bar{s}_{\tau,k}<\rho_r,
\end{cases}
\label{eq:mode-selection}
\end{equation}
where $0<\rho_r<\rho_c<1$. \texttt{EXPAND} adds concrete support when the policy rarely succeeds; \texttt{REVISE} adjusts existing entries as the remaining failures change; \texttt{COMPRESS} removes redundant entries once the policy succeeds reliably; and \texttt{FORCED\_PRUNE} enforces the hard capacity budget. Because $\bar{s}_{\tau,k}$ is task-type-specific, different task types can receive different degrees of support at the same training step. Appendix~\ref{app:mixed-groups} gives the binary-reward observation that mixed success--failure groups, and hence richer contrastive evidence and non-degenerate group-relative returns, occur most often at intermediate success probabilities.

\paragraph{Atomic Commit.}
The refiner proposes edits but never writes the scaffold directly. Its output is limited to four operation types: proposing a principle, proposing a mistake, updating an entry, or deleting an entry. A deterministic validator first checks schema validity and evidence support: new entries require at least two supporting cards, while updates and deletions must reference an existing entry. It then enforces duplicate, mode-specific edit-budget, direction, and global-capacity constraints on the proposed post-edit state. Only a valid proposal is atomically committed; otherwise, the update is logged as a no-op and the snapshot remains unchanged. The modes differ in allowed edit direction: \texttt{EXPAND} permits bounded additions, \texttt{REVISE} prioritizes updates and merges, \texttt{COMPRESS} requires non-increasing scaffold size, and \texttt{FORCED\_PRUNE} permits only reduction.

This design gives \texttt{REVISE} a distinct role. Without it, the controller could only add support or delete it. Revision instead changes the scaffold as the policy's residual failures shift, without assuming that the next useful intervention is either more text or immediate removal. It may narrow, broaden, merge, or replace existing guidance under a fixed edit budget; \texttt{COMPRESS} later reduces support when competence is high. Full JSON schemas, operation budgets, duplicate checks, and commit logs are reported in Appendix~\ref{app:stage1-interface}.

\subsection{Training Objective}
\label{sec:policy-optimization}

\paragraph{Scaffold-Conditioned Warmup.}
The online scheduling mechanism above assumes that the policy can understand the scaffold schema and ground natural-language principles in environment actions. Simply inserting a scaffold block into an unadapted base policy need not establish a stable scaffold interface. Following the cold-start design of skill-augmented RL \citep{skillrl2026}, we independently perform one scaffold-conditioned supervised fine-tuning stage before RL. Given a demonstration set $\mathcal{D}_{\mathrm{SFT}}=\{(x_j,z_j,y_j)\}$, where $z_j$ is a structured scaffold view paired with input $x_j$, the standard supervised objective \citep{ouyang2022} is
\begin{equation}
\mathcal{L}_{\mathrm{SFT}}(\theta)
=
-\mathbb{E}_{(x,z,y)\sim\mathcal{D}_{\mathrm{SFT}}}
\left[
\sum_{t=1}^{|y|}
\log\pi_\theta(y_t\mid x,z,y_{<t})
\right].
\label{eq:sft}
\end{equation}
This stage initializes only the ability to read and apply scaffold entries; it neither generates, filters, nor updates entries online. \method begins scaffold scheduling from the resulting policy parameters, and subsequent changes in scaffold state are driven entirely by on-policy trajectory evidence.

\paragraph{RL-based Policy Optimization.}
The training scaffold changes from trajectories sampled by the current policy, but it is not part of the differentiable policy objective. The policy is optimized only by environmental rewards. For $n$ trajectories of the same task, the group-normalized advantage is
\begin{equation}
\widehat{A}_i
=
\frac{r_i-\bar r}{\sigma_r+\epsilon_{\mathrm{std}}},
\qquad
\bar r=\frac{1}{n}\sum_{j=1}^{n}r_j.
\label{eq:grpo-advantage}
\end{equation}
Here, $\sigma_r$ is the within-group reward standard deviation and $\epsilon_{\mathrm{std}}>0$ prevents division by zero. Let $(y_{i,1},\ldots,y_{i,L_i})$ denote the policy-generated tokens in trajectory $\xi_i$ that participate in optimization, and let $\kappa_{i,\ell}$ be the causal context of token $\ell$, including the task, interaction history, and scaffold. Its importance ratio is
\begin{equation}
\rho_{i,\ell}(\theta)
=
\frac{\pi_\theta(y_{i,\ell}\mid \kappa_{i,\ell})}
{\pi_{\theta_{\mathrm{old}}}(y_{i,\ell}\mid \kappa_{i,\ell})}.
\label{eq:importance-ratio}
\end{equation}
We use the PPO-style clipped GRPO objective \citep{ppo2017,deepseekmath2024}:
\begin{equation}
\begin{aligned}
&\mathcal{L}_{\mathrm{GRPO}}(\theta)
=-\frac{1}{n}\sum_{i=1}^{n}\frac{1}{L_i}\sum_{\ell=1}^{L_i}
\min\Big(
\rho_{i,\ell}(\theta)\widehat A_i,\\
&\operatorname{clip}\!\big(
\rho_{i,\ell}(\theta),1-\epsilon_{\mathrm{clip}},1+\epsilon_{\mathrm{clip}}
\big)\widehat A_i
\Big)
+\beta D_{\mathrm{KL}}(\pi_\theta\|\pi_{\mathrm{ref}}).
\end{aligned}
\label{eq:grpo-objective}
\end{equation}
Here, $\epsilon_{\mathrm{clip}}$ is the clipping radius, $\beta$ is the KL coefficient, and $\pi_{\mathrm{ref}}$ is the fixed reference policy.
Each training iteration uses one scaffold snapshot to generate rollouts. The policy update and evidence construction consume the same trajectories, but only the former produces gradients. The refiner then proposes an edit, deterministic validation checks it, and an atomic commit creates the next snapshot. This snapshot takes effect only in the next iteration. At the end of training, we discard the refiner, evidence builder, scaffold controller, and external scaffold state, and use the final policy with $z=\varnothing$.

\section{Experiments}

In this section, we conduct extensive experiments on ALFWorld, WebShop, and seven search-augmented QA benchmarks to evaluate \method, designed to answer the following research questions:
\begin{itemize}
    \item \textbf{RQ1:} How does \method compare with SOTA baselines in performance and token efficiency?
    \item \textbf{RQ2:} How do the individual components and variants of \method contribute to the overall gains?
    \item \textbf{RQ3:} Does training scaffolding fundamentally change test-time behavior without skills?
\end{itemize}

\subsection{Performance and Efficiency Comparison (RQ1)}

\begin{table*}[t]
\centering
\small
\setlength{\tabcolsep}{3pt}
\begin{tabular}{@{}l*{8}{c}@{\hspace{6pt}}|@{\hspace{6pt}}*{3}{c}@{}}
\toprule
Method &
\multicolumn{8}{c}{ALFWorld} &
\multicolumn{3}{c}{WebShop}\\
\cmidrule(r{12pt}){2-9}
\cmidrule(r{2pt}){10-12}
& Pick & Look & Clean & Heat & Cool & Pick2 & All SR $\uparrow$ & Tok. $\downarrow$ & Score $\uparrow$ & SR $\uparrow$ & Tok. $\downarrow$\\
\midrule
\multicolumn{12}{l}{\emph{Qwen2.5-1.5B-Instruct}}\\
GRPO & 79.0 & \textbf{69.2} & 71.6 & 72.9 & 62.7 & 48.6 & $67.86\pm3.50$ & 14,723 & 0.731 & $52.80\pm1.66$ & 13,451\\
SKILL0 & 71.4 & 43.6 & 61.7 & 62.5 & 56.0 & 38.9 & $57.62\pm3.21$ & 16,581 & 0.504 & $30.80\pm8.35$ & 29,839\\
SkillRL & 79.0 & 41.0 & 16.0 & 52.1 & 37.3 & 47.2 & $47.38\pm9.44$ & 17,749 & 0.699 & $41.80\pm13.11$ & 16,640\\
SkillRL$^\dagger$ & 86.7 & 43.6 & \textbf{86.4} & 64.6 & 77.3 & 48.6 & $71.90\pm2.63$ & 45,595 & 0.575 & $29.07\pm23.09$ & 23,144\\
\method$^\dagger$ & 89.5 & 48.7 & 81.5 & \textbf{83.3} & 77.3 & 76.4 & $79.05\pm0.34$ & 12,901 & 0.783 & $\mathbf{59.47\pm2.47}$ & 10,798\\
\textbf{\method} & \textbf{91.4} & 51.3 & 85.2 & 81.2 & \textbf{78.7} & \textbf{77.8} & $\mathbf{80.71\pm1.54}$ & \textbf{10,245} & \textbf{0.795} & $56.33\pm7.06$ & \textbf{9,184}\\
\midrule
\multicolumn{12}{l}{\emph{Qwen2.5-7B-Instruct}}\\
GRPO & 89.5 & 66.7 & 74.1 & 64.6 & 61.3 & 61.1 & $71.67\pm2.99$ & 17,060 & 0.784 & $57.60\pm10.33$ & 12,587\\
RLOO & 92.4 & 61.5 & 90.1 & 68.8 & 72.0 & 65.3 & $78.10\pm4.30$ & 14,676 & 0.852 & $72.73\pm2.72$ & 13,013\\
SKILL0 & 88.6 & 66.7 & 86.4 & 81.2 & 66.7 & 69.4 & $78.10\pm5.03$ & 14,162 & 0.789 & $63.27\pm0.74$ & 15,078\\
SkillRL & \textbf{98.1} & 76.9 & 91.4 & 91.7 & 58.7 & 84.7 & $84.76\pm1.23$ & 11,908 & 0.858 & $70.07\pm5.26$ & 8,077\\
SkillRL$^\dagger$ & \textbf{98.1} & 74.4 & \textbf{97.5} & 83.3 & \textbf{92.0} & \textbf{88.9} & $\mathbf{91.43\pm2.10}$ & 29,933 & 0.835 & $72.07\pm1.79$ & 17,763\\
\method$^\dagger$ & 94.3 & \textbf{82.1} & \textbf{97.5} & \textbf{93.8} & 88.0 & 79.2 & $90.00\pm2.33$ & 11,774 & \textbf{0.900} & $78.20\pm0.20$ & 7,881\\
\textbf{\method} & 93.3 & 76.9 & 96.3 & 89.6 & 89.3 & 81.9 & $89.29\pm1.01$ & \textbf{10,455} & 0.898 & $\mathbf{78.40\pm0.20}$ & \textbf{7,749}\\
\bottomrule
\end{tabular}
\caption{Performance on ALFWorld and WebShop. Success rate (SR) values are reported as three-seed means $\pm$ standard deviation. $\dagger$ indicates methods that retain external skills at test time. Bold text highlights \method without external skills. Complete results for 7B models are provided in the Appendix~\ref{app:prompting-baselines}.}
\label{tab:main-results}
\end{table*}

\paragraph{Benchmarks and metrics.}
We evaluate \method on ALFWorld \citep{alfworld2020}, WebShop \citep{webshop2022} and search-augmented QA on seven single- and multi-hop benchmarks. Notably, unlike prior works that rely on inconsistent evaluation sub
set, we conduct all evaluations on the full, fixed test sets of all benchmarks to eliminate selection bias and enasure strict comparability. More evaluation details are in Appendix~\ref{app:stage1-interface}; the Search protocol is in Appendix~\ref{app:search-qa}.

\paragraph{Experimental Settings}
For ALFWorld and WebShop, we train \mbox{Qwen2.5-1.5B-Instruct} and \mbox{Qwen2.5-7B-Instruct} \citep{qwen25technicalreport2024}. Unless specified otherwise in ablations, \method first undergoes scaffold-conditioned warmup and then executes 150 GRPO steps with eight rollouts per task. The complete Stage~1 training setup is provided in Appendix~\ref{app:stage1-interface}. Search-augmented QA experiments are conducted using the 7B model following the protocol in Appendix~\ref{app:search-qa}.

\paragraph{Results on ALFWorld and WebShop.}
Table~\ref{tab:main-results} compares \method with native baselines under identical training and evaluation setups. We report SkillRL with and without test-time skills; unless specified as a diagnostic ("supported"), \method strictly operates in the skill-free regime. Note that ``Avg. tokens'' accounts for the total deployment context (prompt + response) during inference, rather than tokens used for loss computation during training. Across 1.5B/7B, \method improves over GRPO by 12.9/17.6 ALFWorld SR points and 3.5/20.8 WebShop SR points, raises WebShop Score by 0.064/0.114, and reduces deployment interaction tokens by 30.4/38.7\% on ALFWorld and 31.7/38.4\% on WebShop. Removing the scaffold changes ALFWorld SR by $+1.66/-0.71$ points and WebShop Score by at most 0.012, supporting scaffold-free deployment without claiming that every entry has zero marginal value. Appendix Table~\ref{tab:web-categories} reports WebShop categories. Across all four model--environment pairs, scaffold-free \method exceeds same-scale GRPO, confirming that the gains survive support removal.


\paragraph{Results on Search-augmented QA Tasks.}
We further evaluate 7B policies on seven search-augmented QA benchmarks, training on NQ and HotpotQA~\citep{naturalquestions2019,hotpotqa2018} and testing transfer to five out-of-domain datasets. Table~\ref{tab:search-main} compares \method with Search-R1, ZeroSearch, EvolveR, SKILL0, and SkillRL~\citep{searchr12025,zerosearch2025,evolver2025,skillzero2026,skillrl2026}. Appendix~\ref{app:search-qa} gives the retrieval protocol, dataset citations, and per-dataset results. \method improves over SKILL0 by 3.5 average points with 7.6\% more prompt tokens. SkillRL has the highest reported average but retains its skill library; \method uses 32.1\% fewer prompt tokens after removing its training scaffold. The comparison therefore measures a performance--context trade-off after scaffold removal.

\begin{table}[h]
\centering
\small
\setlength{\tabcolsep}{5pt}
\begin{tabular}{lcc}
\toprule
Method & Avg. $\uparrow$ & Avg. Prompt Tok. $\downarrow$\\
\midrule
Search-R1$^\ddagger$ & 38.5 & --\\
ZeroSearch$^\ddagger$ & 39.1 & --\\
EvolveR$^\ddagger$ & 43.1 & --\\
SKILL0 & 41.7 & \textbf{688}\\
SkillRL$^{\dagger}$ & \textbf{46.0} & 1,090.1\\
\textbf{\method} & 45.2 & 740\\
\bottomrule
\end{tabular}
\caption{Experimental results on search-augmented QA benchmarks. \method is evaluated without test-time skills; $\dagger$ indicates methods that retain  external skills. $\ddagger$ denotes results reported by SkillRL. }
\label{tab:search-main}
\vspace{-1em}
\end{table}

\subsection{Ablation Study and Mechanistic Analysis(RQ2)}

With interface initialization fixed, removing the online scaffold costs 7.14 SR points on 1.5B ALFWorld (Table~\ref{tab:ablation}), apart from the end-to-end gain over GRPO. All online variants underperform the full method; removing semantic revision causes the largest controller drop ($-12.62$), supporting its role between detailed guidance and deletion. The trajectory-wise variant is also substantially more variable, consistent with group aggregation reducing sensitivity to individual rollout noise. The frozen and \texttt{EXPAND}-only variants further separate adaptation from accumulation: retaining or adding context is insufficient when its content cannot track the policy's changing residual failures. The warm-start alternative and interface-removal control separately test Bank initialization and scaffold-interface SFT; Appendix~\ref{app:stage1-interface} reports the dynamics. Independent 7B WebShop ablations show the same broad pattern: online adaptation and group-level review help, while expansion-only editing is highly unstable (Appendix~\ref{app:webshop-ablation}).

\begin{table}[h]
\centering
\small
\setlength{\tabcolsep}{3.5pt}
\begin{tabular}{lcc}
\toprule
Variant & SR (\%) & $\Delta$SR\\
\midrule
\textbf{\method} & $\mathbf{80.71\pm1.54}$ & 0.00\\
w/o training scaffold & $73.57\pm0.89$ & $-7.14$\\
Frozen after step 50 & $77.14\pm4.77$ & $-3.57$\\
Trajectory-wise evidence & $74.05\pm9.04$ & $-6.66$\\
\texttt{EXPAND} only & $72.86\pm8.81$ & $-7.85$\\
w/o \texttt{REVISE} & $68.10\pm6.45$ & $-12.62$\\
SkillRL-style warm start & $70.24\pm7.73$ & $-10.47$\\
w/o interface initialization & $47.86\pm3.29$ & $-32.85$\\
Earlier architecture + warm start & $43.57\pm2.40$ & $-37.14$\\
\bottomrule
\end{tabular}
\caption{Ablation study of the 1.5B model on ALFWorld, with all variants evaluated without skills. The full \method and five online variants share the same warmup checkpoint. Further details are reported in the Appendix~\ref{app:offline-scaffold-replay}.}
\label{tab:ablation}
\vspace{-1em}
\end{table}

\paragraph{Inference-Effective Skills vs.\ Training-Effective Skills.}
Offline replay reveals a gap between inference utility and training utility. Across five fixed checkpoints, the static 55-entry library initially improves success and behavioral coverage, but its contribution to within-group reward variation diminishes as competence rises. A clean-and-place case similarly shows that policy-derived support can produce more outcome-balanced exploration despite lower immediate success. Consistently, the static-bank warm start reaches $70.24\pm7.73$ unsupported SR, 10.47 points below the empty \method Bank.  Thus, inference-effective skills need not provide the most useful training support; Appendix~\ref{app:offline-scaffold-replay} reports the complete statistics and trajectories.

Across three 1.5B ALFWorld runs, controller reviews shift from early \texttt{EXPAND} to \texttt{REVISE} and later \texttt{COMPRESS}, while \texttt{FORCED\_PRUNE} is rare. This is not a fixed withdrawal schedule: semantic revision dominates between early acquisition and late compression, allowing the Bank to track residual failures before support is removed. This operation sequence explains the non-monotonic context curve in Figure~\ref{fig:alfworld-training-dynamics}: expansion increases early support, revision changes its content without requiring continued growth, and compression produces the later contraction. Appendix~\ref{app:stage1-interface} reports the dynamics, and Appendices~\ref{app:scaffold-evolution}--\ref{app:cross-stage-audit} audit the edits.

\subsection{Analyze behavioral shifts induced by training scaffolds (RQ3)}
\label{sec:stage0-diagnostic}

we analyze how training scaffold alters the intrinsic policy behavior when test-time skills are removed. On ALFWorld with Qwen2.5-1.5B-Instruct, we construct the fixed natural-language skills from training-task successes and failures. Starting from same checkpoint, \emph{skill-conditioned RL} receives thess skills during GRPO and \emph{skill-free RL} does not. Appendix~\ref{app:stage0-details} gives the detail evaluation settings.

As evaluation targets, we collect 24 successful trajectories generated by the initial model under skill guidance. To assess intrinsic policy shifts without test-time hints, we compute the skill-free teacher-forced NLL of both trained models on these target responses. Specifically, for each fixed state $x_j$, skill $b$, and target response $y_j^b$, the skill-free teacher-forced NLL over the full sequence is defined as:
\begin{equation}
\mathcal L_{\mathrm{full}}^{\mathrm{free}}(\theta) = -\frac{1}{|\mathcal I|}\sum_{(j,l)\in\mathcal I} \log\pi_\theta(y_{j,l}^b\mid x_j, y_{j,<l}^b),
\end{equation}
alongside its supported counterpart $\mathcal L_{\mathrm{full}}^{\mathrm{skill}}(\theta)$ where skill $b$ is present in the prompt. We then define the \emph{NLL Gap} as $\Delta \text{NLL}(\theta) = \mathcal L_{\mathrm{full}}^{\mathrm{free}}(\theta) - \mathcal L_{\mathrm{full}}^{\mathrm{skill}}(\theta)$ which quantifies the marginal likelihood gain contributed by skill-conditioned RL traiing on the exact same response tokens.

\begin{figure}[h]
\centering
\includegraphics[width=0.95\columnwidth]{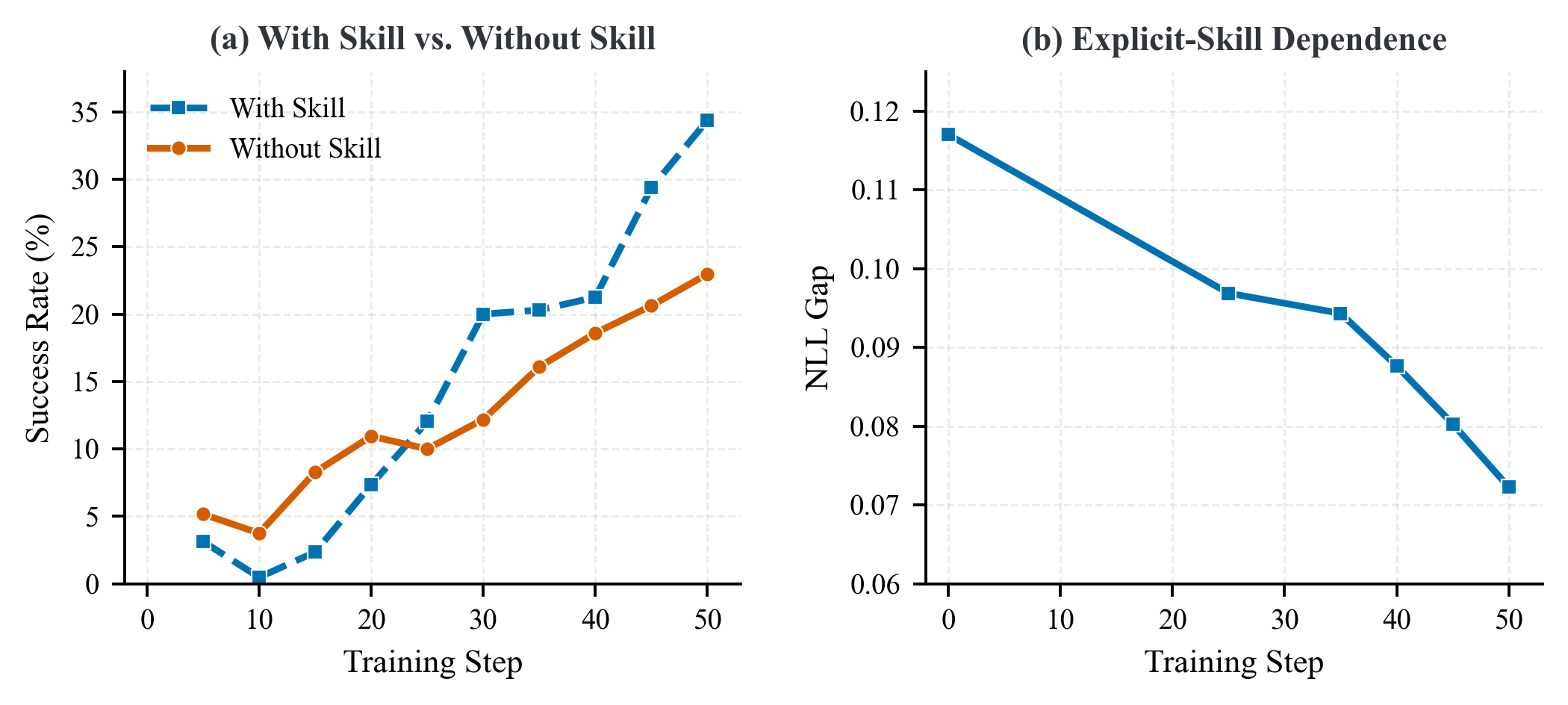}
\caption{NLL Gap between two variants. The narrowing NLL gap  demonstrates that explicit guidance is gradually parameterised into the model parameters.}
\label{fig:stage0}
\end{figure}

After 50 steps, the NLL gap drops by a relative 38.3\% (Figure~\ref{fig:stage0}), indicating that the model progressively internalizes the skill knowledge. Consequently, when evaluated on skill-free tasks, the scaffold-trained model significantly outperforms the baseline model trained without scaffolds ($\theta_{\mathrm{no\text{-}scaffold}}$), achieving a lower NLL (0.6014 vs.\ 0.6356) and a higher success rate ($22.97\%\pm6.39\%$ vs.\ $17.66\%\pm7.94\%$). Furthermore, reintroducing skills at test time boosts performance to $34.38\%\pm7.35\%$. Importantly, because the benchmark spans only two gamefiles, these results demonstrate controlled skill internalization for this specific distribution rather than general skill acquisition.

\section{Conclusion}
In this work, we introduced \method, a policy-aware training scaffold that adaptively adjusts skill-based context to match the needs of the policy across different learning stages. By converting recent rollout trajectory groups into evidence cards and updating guidance via task-specific evaluations, \method delivers targeted guidance when the policy is weak and systematically redacts redundant context as capabilities mature. Our findings reveal a fundamental insight: skill optimized for inference is not necessarily the most effective training signal. Instead, optimal training support must actively preserve informative rollout contrast to drive RL exploration. Across extensive evaluations on ALFWorld, WebShop, and seven search-augmented QA benchmarks, \method achieves performance competitive with state-of-the-art baselines while reducing total token consumption by 25\%--50\%.

\bibliography{references}

\clearpage
\onecolumn
\appendix
\section*{Supplementary Material for
PATS: Policy-Aware Training Scaffolding for Agentic Reinforcement Learning}

\section{Stage 0 Controlled-Diagnostic Details}
\label{app:stage0-details}

\subsection{Training and Evaluation Configuration}

The Stage-0 study uses Qwen2.5-1.5B-Instruct and ALFWorld. We draw 16 training tasks, balanced across the six task types, from the training split. We reserve a further 16 non-overlapping tasks for held-out behavioral evaluation. Each condition uses five random seeds and samples eight trajectories per task. Sampling uses temperature 0.4, top-$p=0.95$, and at most 512 new tokens, with decoding stopped at \texttt{</action>}. GRPO uses learning rate $10^{-6}$, train batch size 16, PPO mini-batch size 128, rollout group size 8, KL coefficient 0.01, maximum prompt length 2048, maximum response length 512, an environment limit of 30 steps, and history length two.

\subsection{Skill Band, Controls, and NLL Set}
Stage~0 first constructs a system-message skill band from empty experience text while keeping model parameters frozen. The historically best band occurs in round four, achieves 18.75\% success on the training tasks, and contains ten natural-language rules totaling approximately 737 characters. The baseline injects no explicit band, thereby controlling for the effect of ordinary RL updating on behavior after band removal.

The NLL evaluation does not re-roll out the environment. Instead, it freezes 24 oracle trajectories sampled from successful trajectories at the best first-stage iteration as $\mathcal{D}_{\mathrm{eval}}$. These trajectories comprise 610 environment steps from two \texttt{pick\_and\_place\_simple} gamefiles, and 16 trajectories originate from the same CreditCard--Shelf instance. Full-response NLL scores the entire response, including \texttt{<think>}, \texttt{<action>}, and their boundary tags. Action-only NLL scores only the action content inside \texttt{<action>...</action>}, excluding the tags themselves.

The full-response scoring equations are defined in Section~\ref{sec:stage0-diagnostic}; this appendix fixes the sampling set and clarifies that action-only NLL is used only as an auxiliary audit, not as a training signal or headline diagnostic.

\subsection{Interpretation Boundary}
Behavioral success rates are aggregated over five seeds. Because Stage~0 has only 16 held-out tasks and its NLL set covers only two gamefiles, we use success rate as behavioral corroboration for a likelihood diagnostic rather than as a final benchmark-performance claim. All Phase-C supported evaluations inject the best band; consequently, supported results in the random group include a train/evaluation band mismatch. The main text therefore emphasizes the no-skill control comparison. The diagnostic complements, rather than replaces, the Stage~1 benchmark: it tests a fixed response distribution under controlled removal of one band and is not used as reward, advantage, or a controller signal.

\section{Stage 1 Details and Supplementary Results}
\label{app:stage1-interface}

\subsection{Training and Evaluation Configuration}

\begin{center}

\begin{minipage}[c]{0.40\textwidth}
\centering
\small
\setlength{\tabcolsep}{4pt}
\begin{tabular}{lrr}
\toprule
Configuration & 1.5B & 7B\\
\midrule
RL algorithm & GRPO & GRPO\\
Training steps & 150 & 150\\
Rollout group size & 8 & 8\\
Train batch size & 16 & 16\\
PPO mini-batch size & 256 & 256\\
PPO micro-batch/GPU & 4 & 2\\
Learning rate & $10^{-6}$ & $10^{-6}$\\
Max. environment steps & 50 & 50\\
History length & 2 & 2\\
Number of GPUs & 8 & 8\\
Training seeds & 0,1,2 & 0,1,2\\
\bottomrule
\end{tabular}

\captionof{table}{Shared Stage-1 RL configuration.}
\label{tab:rl-config}
\end{minipage}
\hfill
\begin{minipage}[c]{0.52\textwidth}
\centering
\includegraphics[
    width=0.92\linewidth
]{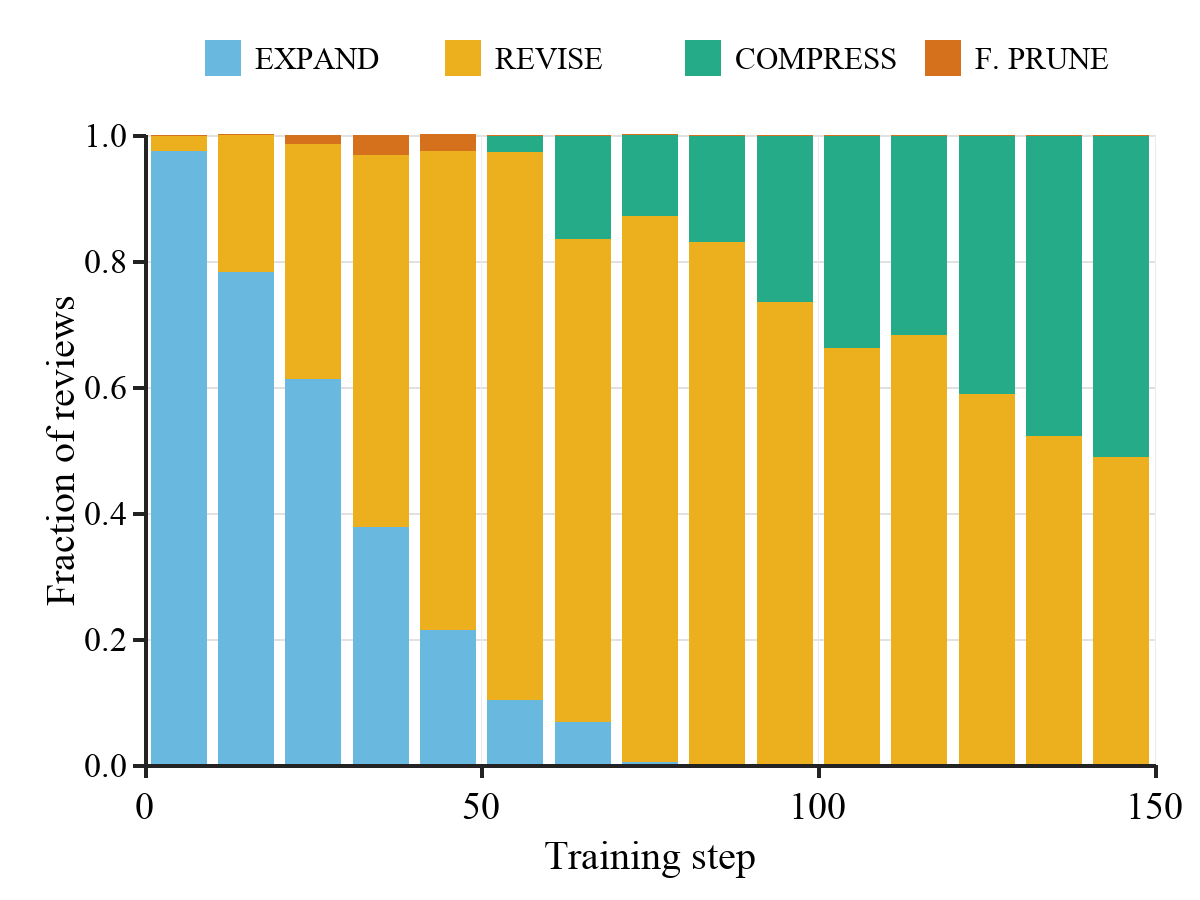}

\captionof{figure}{Controller-mode evolution on 1.5B ALFWorld.}
\label{fig:controller-modes}
\end{minipage}

\par\medskip

\includegraphics[
    width=0.76\textwidth
]{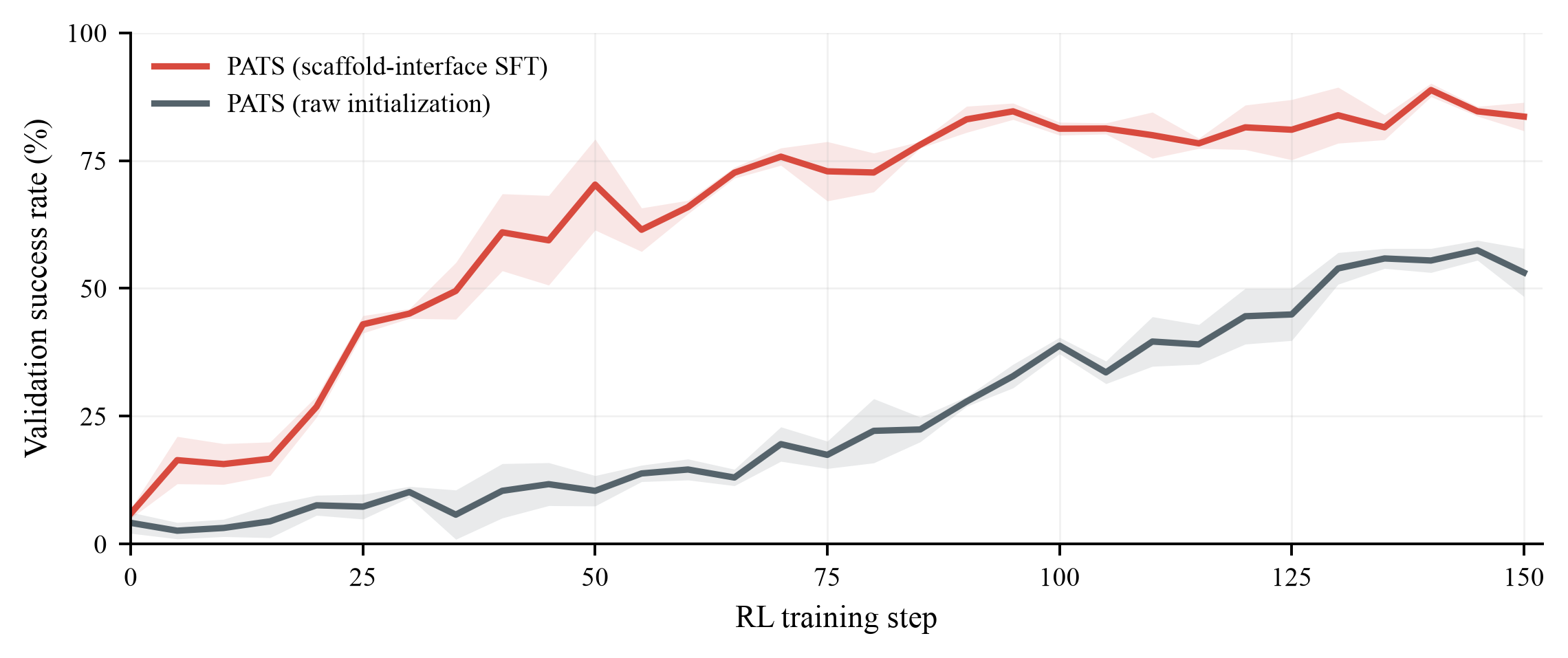}

\captionof{figure}{
Interface initialization precedes Stage~1.
Step-zero validation differs only modestly, but the initialized policy
learns much faster once both conditions begin online RL with dynamic support.
}
\label{fig:sft-dynamics}

\end{center}

Evaluation uses the final step-150 checkpoint, one trajectory per task, temperature 0.4, top-$p=1.0$, and at most 50 environment steps. ALFWorld uses its fixed 140 tasks; WebShop uses the canonical 500-goal set (\texttt{seed=0}, \texttt{env\_seed=1000}, and \texttt{goal\_idxs=range(500)}). All methods share environments, sampling parameters, and success criteria, but retain their native prompt renderers rather than evaluating another method's checkpoint with the \method prompt format. Policy prompts and responses are limited to 3072 and 512 tokens. Training uses KL coefficient 0.01, $\gamma=0.95$, invalid-action penalty 0.1, validation every five steps, checkpointing every ten steps, and one validation pass before training.

The scaffold-interface SFT follows the public SkillRL ALFWorld data configuration \citep{skillrl2026}. The formal 1.5B run initializes from Qwen2.5-1.5B-Instruct and trains for three epochs on eight GPUs (27 steps per epoch; 81 steps total), using \texttt{instruction} as input and \texttt{output} as target. It uses maximum length 2048, learning rate $10^{-5}$, global batch size 256, per-GPU micro-batch 4, cosine decay, 10\% warmup, weight decay 0.01, gradient clipping 1.0, and the final global-step-81 checkpoint. This public data initializes a scaffold interface; it is not a new distillation dataset proposed in this work. Early-RL validation uses the training logs to distinguish interface initialization from subsequent on-policy scaffold adaptation.

Online scaffold adaptation uses an independent \texttt{qwen2.5-7b-refiner} service. After the policy update at each RL iteration, it reads evidence cards and the current committed scaffold state for each represented task type, then returns JSON tool calls. It never generates agent actions and is not part of the GRPO objective. Service failures, timeouts, JSON parsing errors, and schema-invalid proposals are logged; the corresponding edit is skipped without interrupting policy training. The service, evidence builder, and controller exist only during training and are removed at deployment.

One audited 1.5B ALFWorld run uses eight H20 GPUs for approximately 22 hours 26 minutes (about 180 GPU-hours). Its seed-0 log records 831 refiner requests, 801 successful calls, 44.1M input characters, 0.53M output characters, and 3.2 seconds mean latency. These are training-time service costs, not deployment tokens.

\subsection{Policy Prompt, Evidence Card, and Refiner Interface}

At each policy call, the original task instruction is followed by a \texttt{Retrieved Relevant Experience} block and then the current history and observation. Non-empty sections render general principles, type-specific skills, and mistakes. Each skill contains a short title, an actionable principle, and an ``apply when'' condition; each mistake pairs a prohibited pattern with a corrective action. Unsupported evaluation removes the entire experience block while retaining the original task renderer, history, observation, action format, and decoding settings.

The implementation logs and prompts use \emph{bank} or \emph{SkillBank} as code-level names for the same scaffold state described in the main paper. The deterministic evidence builder records success/failure counts, trajectory length, invalid actions, recurring action/observation patterns, grounded environment signals, and representative successful and failed traces. It normalizes actions as environment commands and excludes policy reasoning text. The refiner also receives training step, scaffold version, entry/token pressure, task-type success EMA, selected mode, relevant entries, and a title-only digest of other task types. Its output is restricted to propose-skill, propose-mistake, update-skill, and delete-skill objects. New entries require two supporting evidence cards and pass length, reference, duplication, mode-budget, and capacity checks.

\subsubsection{Policy-Facing Experience Block}
The following block is inserted between the original task introduction and the current interaction history. Unsupported evaluation removes the entire block while leaving the task renderer, history, observation, action format, and decoding settings unchanged.
\begin{figure}[htbp]
\centering
\includegraphics[width=0.94\textwidth]{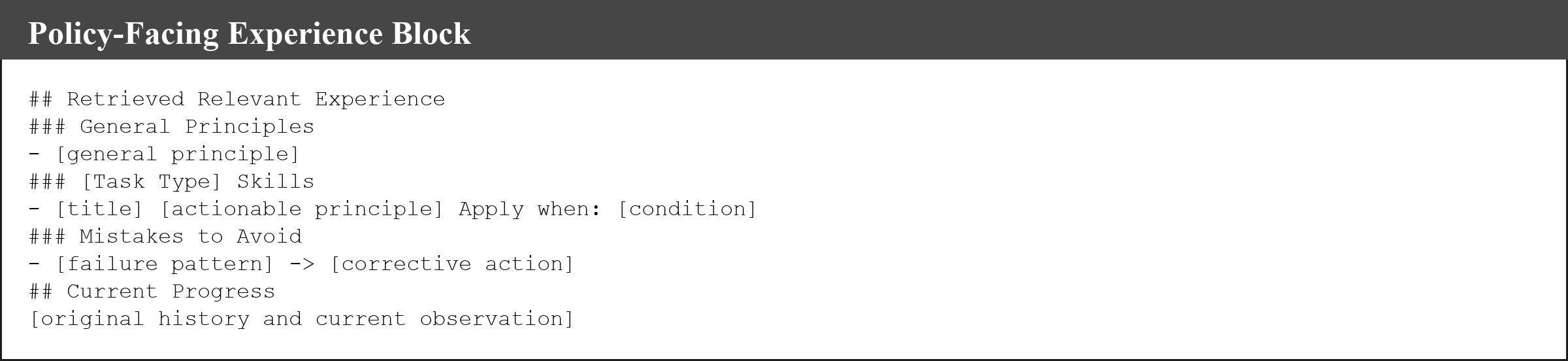}
\caption{Policy-facing experience block rendered before the current interaction history.}
\label{fig:prompt-policy-block}
\end{figure}

\subsubsection{Group-Level Evidence Card}
Each card is constructed deterministically from one rollout group whose trajectories attempt the same concrete task. It contains group success and failure counts, mean trajectory length, invalid-action statistics, recurrent action/observation failure patterns, grounded action--observation contexts that trigger those patterns, and compact representative successful and failed environment traces. Cards are aggregated by task type before scaffold review. Card construction never calls the refiner. Representative traces contain only environment observations and executable actions, and exclude the policy's \texttt{<think>} text.
\begin{figure}[htbp]
\centering
\includegraphics[width=0.94\textwidth]{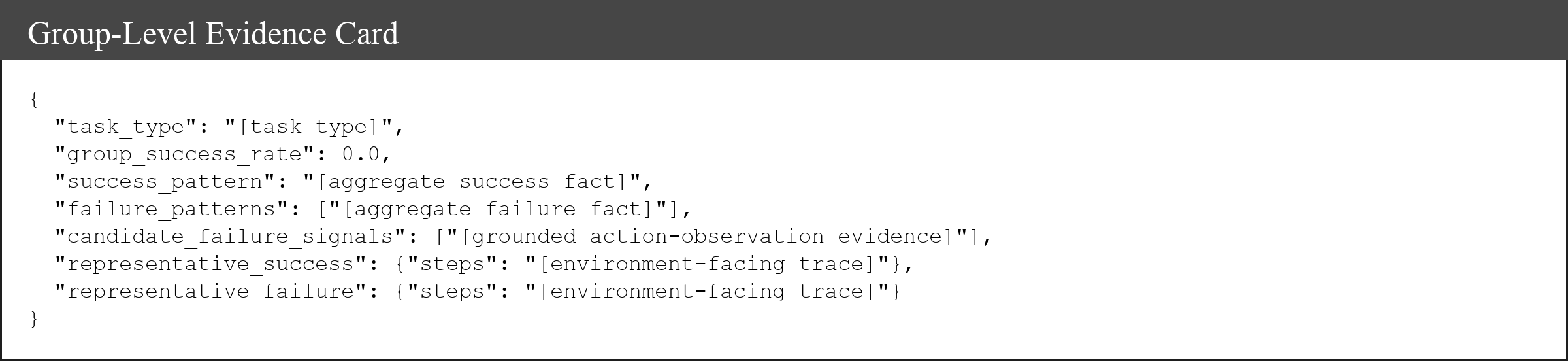}
\caption{Schema of the deterministic group-level evidence card.}
\label{fig:prompt-evidence-card}
\end{figure}

\subsubsection{Refiner Instructions and Constrained Output}
The refiner receives the controller state, the relevant entries, and recent evidence cards. It may propose only \texttt{propose-skill}, \texttt{propose-mistake}, \texttt{update-skill}, or \texttt{delete-skill} objects. \texttt{EXPAND} permits additions; \texttt{REVISE} permits bounded updates; \texttt{COMPRESS} requires an addition to be paired with reduction; \texttt{FORCED\_PRUNE} prohibits additions. Deterministic validation checks schema, references, evidence count, duplication, budget, and capacity before atomic commit.
\begin{figure}[htbp]
\centering
\includegraphics[width=0.94\textwidth]{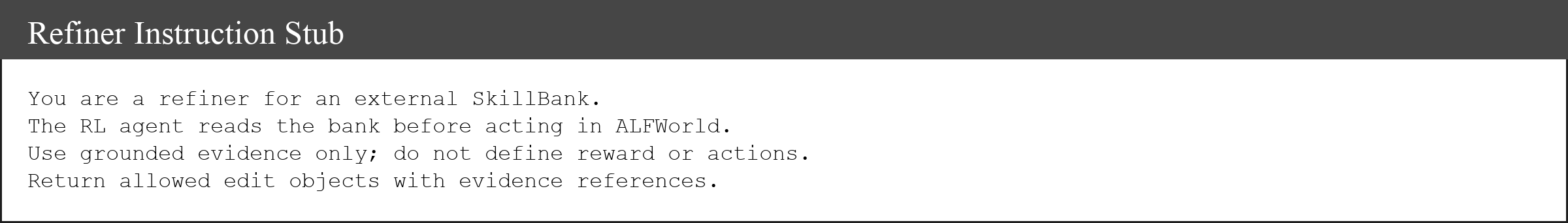}
\caption{Minimal refiner instruction used before the full scaffold review template.}
\label{fig:prompt-refiner-stub}
\end{figure}

\subsubsection{Verbatim Policy Injection Template}
The following template gives the exact text added by \method and its position. The original task introduction, history, and current observation/action instruction are supplied by the ALFWorld renderer and change at each environment step. All headings, field names, punctuation, and entry formatting below match the formal run. The unsupported condition omits the entire experience block. Empty subsections are not rendered. At initialization, when there is no history, the experience block appears before ``Now it's your turn to take an action.'' and no current-progress heading is emitted. The policy always receives one user message; bank version, utility, controller mode, evidence, and refiner metadata are never exposed to it.
\begin{figure}[htbp]
\centering
\includegraphics[width=0.94\textwidth]{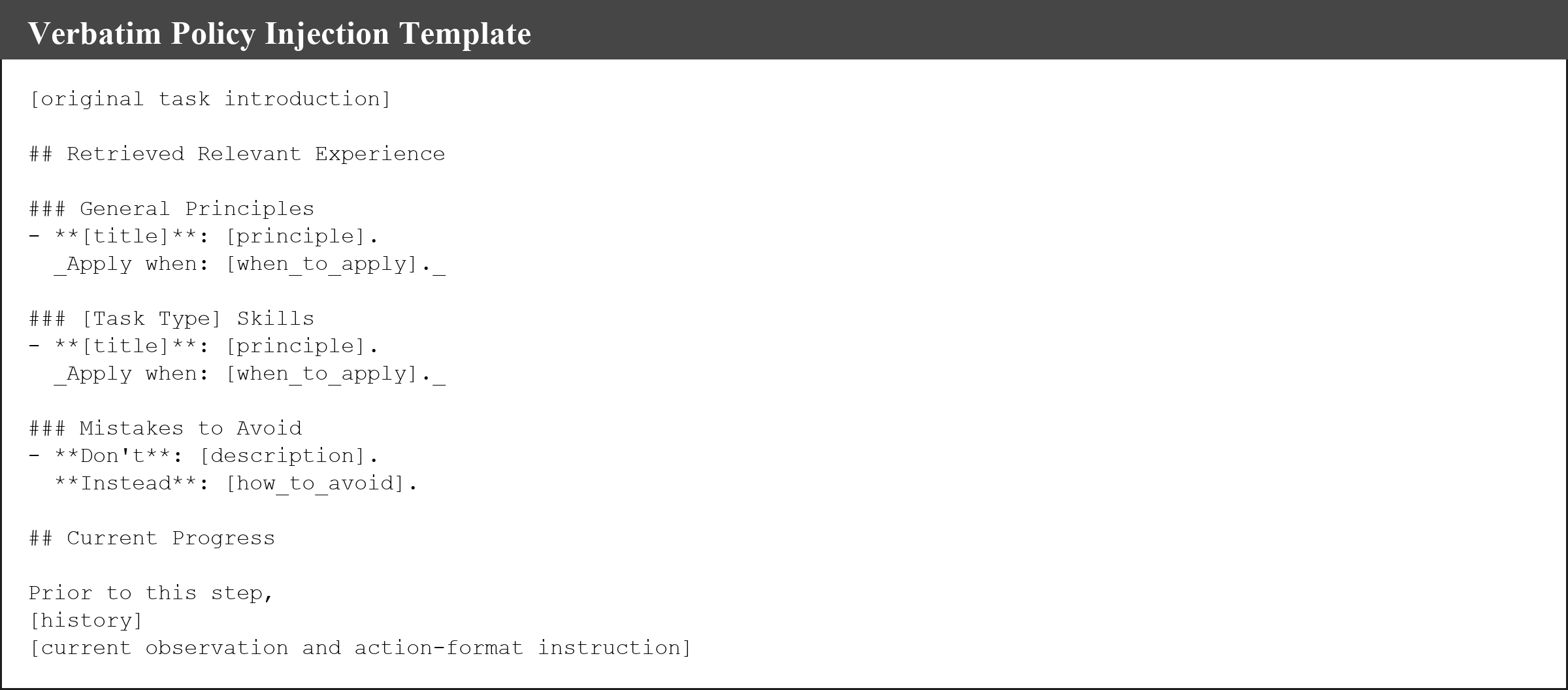}
\caption{Full policy injection template used in the supported training condition.}
\label{fig:prompt-policy-template}
\end{figure}

General principles, task-type skills, and mistakes are drawn from the three layers of the same frozen snapshot. An entry's \texttt{when\_to\_apply} field is a semantic applicability condition, not a literal match against the observation string. The unsupported condition therefore differs only by the absence of this experience block.

\subsubsection{Full Refiner Message and Mode Instructions}
The refiner first receives the following fixed fields, followed by the mode-specific instruction, current frozen scaffold view, and at most 32 evidence cards. The cross-task digest retains titles only: it lets the refiner detect cross-type near-duplicates and promote them to general skills without repeatedly injecting complete entries into the policy.
\begin{figure}[htbp]
\centering
\includegraphics[width=0.94\textwidth]{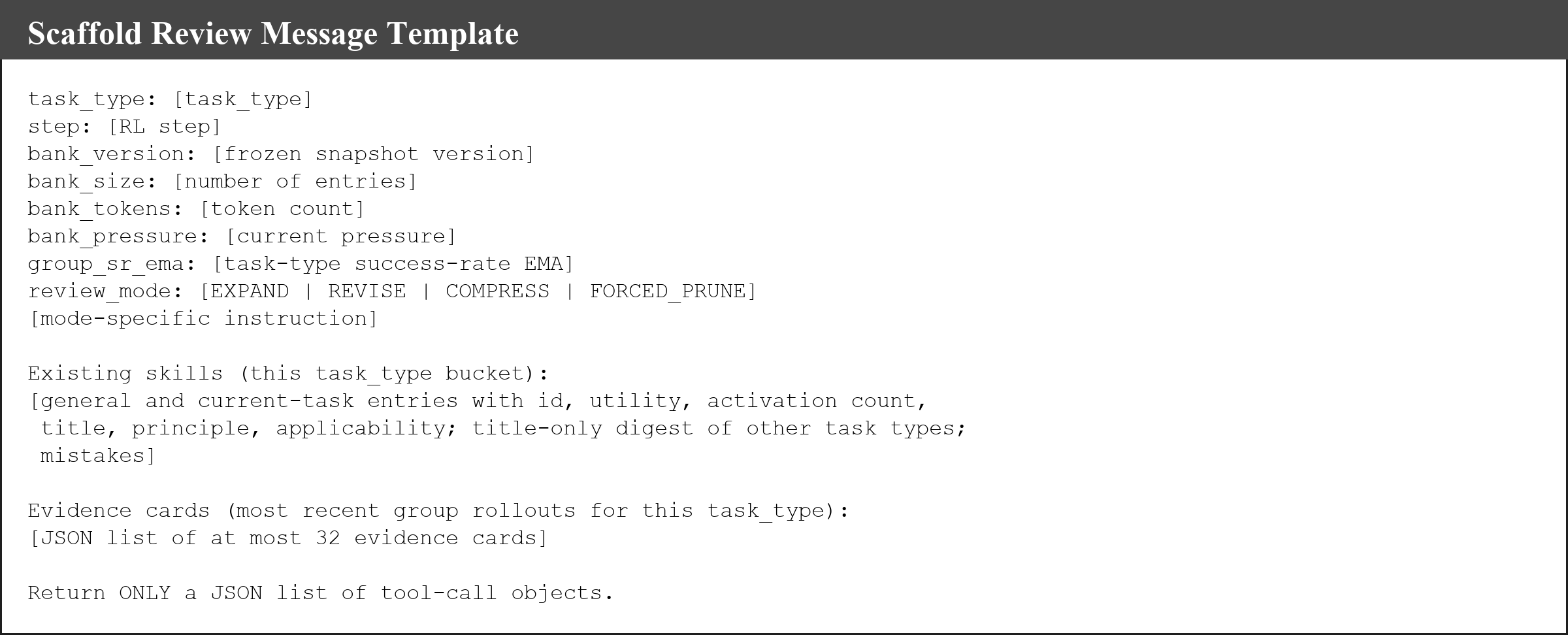}
\caption{Scaffold review message template supplied to the refiner.}
\label{fig:prompt-refiner-message}
\end{figure}
The four mode instructions are:
\begin{figure}[htbp]
\centering
\includegraphics[width=0.94\textwidth]{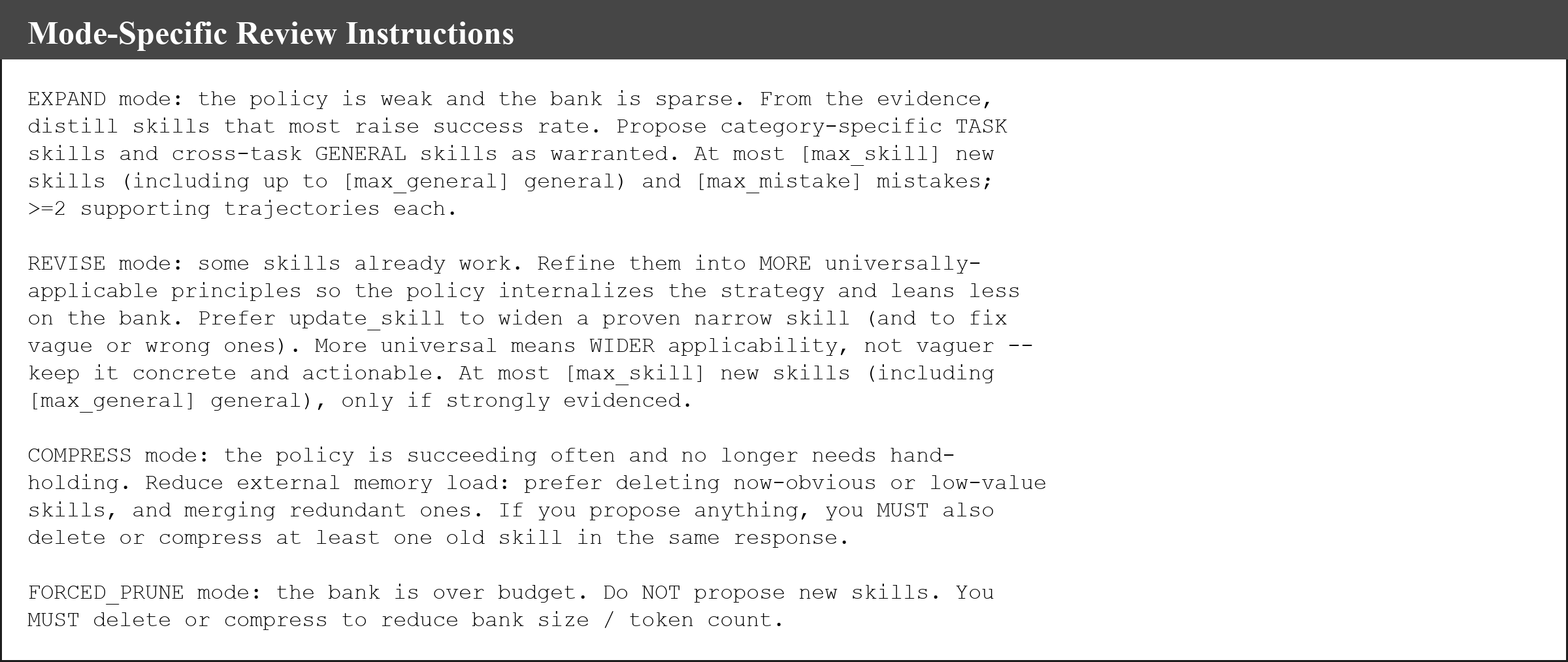}
\caption{Mode-specific scaffold review instructions.}
\label{fig:prompt-mode-instructions}
\end{figure}
Representative traces in evidence cards retain environment observation--action pairs only, never the policy's \texttt{<think>} text. \texttt{EXPAND} distills concrete principles from at least two pieces of evidence; \texttt{REVISE} widens validated narrow rules while keeping them actionable; \texttt{COMPRESS} deletes or merges obvious, low-value, or redundant entries; and \texttt{FORCED\_PRUNE} permits only reduction. Every proposal passes mode-specific budget, schema, duplication, and capacity validation before its post-policy-update atomic commit; the rollout that produced an evidence card always reads the preceding frozen snapshot.

\begin{table}[htbp]
\centering
\small
\setlength{\tabcolsep}{5pt}
\begin{tabular}{lcccp{0.36\textwidth}}
\toprule
Mode & \texttt{max\_skill} & \texttt{max\_general} & \texttt{max\_mistake} & Additional constraint\\
\midrule
\texttt{EXPAND} & 2 & 1 & 1 & Bounded addition for weak policies\\
\texttt{REVISE} & 1 & 1 & 1 & Prefer updates to existing entries\\
\texttt{COMPRESS} & 1 & 1 & 0 & \texttt{max\_total\_propose}=1; must pair with deletion or merging\\
\texttt{FORCED\_PRUNE} & 0 & 0 & 0 & No proposals; reduce scaffold size or tokens\\
\bottomrule
\end{tabular}
\caption{Resolved operation budgets injected into the scaffold review prompt and enforced by deterministic validation.}
\label{tab:mode-budgets}
\end{table}

The macro-review system instruction further constrains the refiner to candidate editing rather than action generation. It distinguishes general skills, task skills, and mistakes; requires a 3--6 word title, a one- or two-sentence principle, and a semantic rather than copied-observation applicability condition; requires at least two evidence cards for each proposed entry; and enforces maximum lengths of 80, 300, and 200 characters for the title, principle, and applicability condition. It directs the refiner to update an existing idea rather than add a paraphrase, retain concrete action verbs, ordering, and required appliances, and generalize interchangeable receptacles and instance IDs while retaining functional appliances such as a microwave for heating, refrigerator for cooling, sink basin for cleaning, and desk lamp for looking. The permitted JSON calls are exactly:
\begin{figure}[htbp]
\centering
\includegraphics[width=0.94\textwidth]{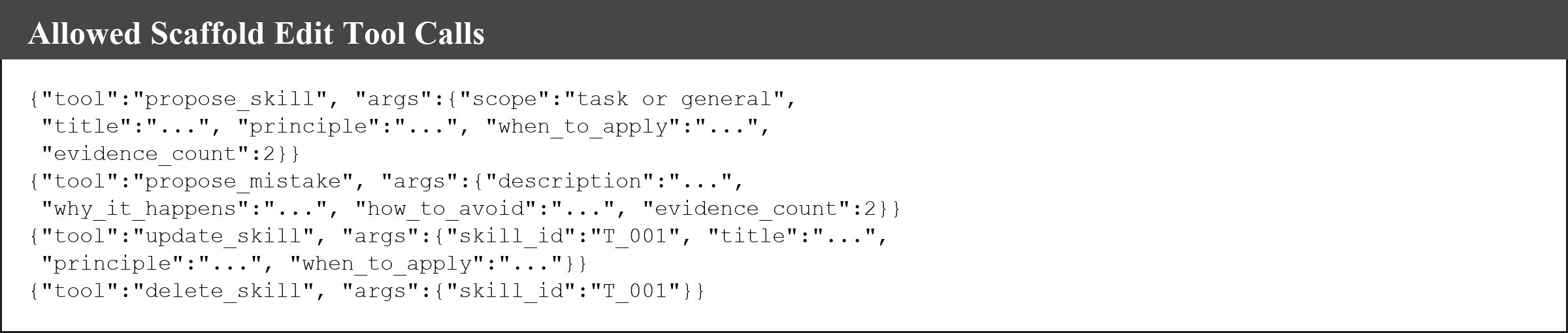}
\caption{Allowed scaffold edit tool calls returned by the refiner.}
\label{fig:prompt-tool-schema}
\end{figure}

For completeness, the formal \texttt{macro\_review} system message is reproduced verbatim below. It limits the language model to proposing candidate edits; all outputs remain subject to code-level budget, schema, deduplication, and capacity validation.
\begin{figure}[htbp]
\centering
\includegraphics[width=0.94\textwidth]{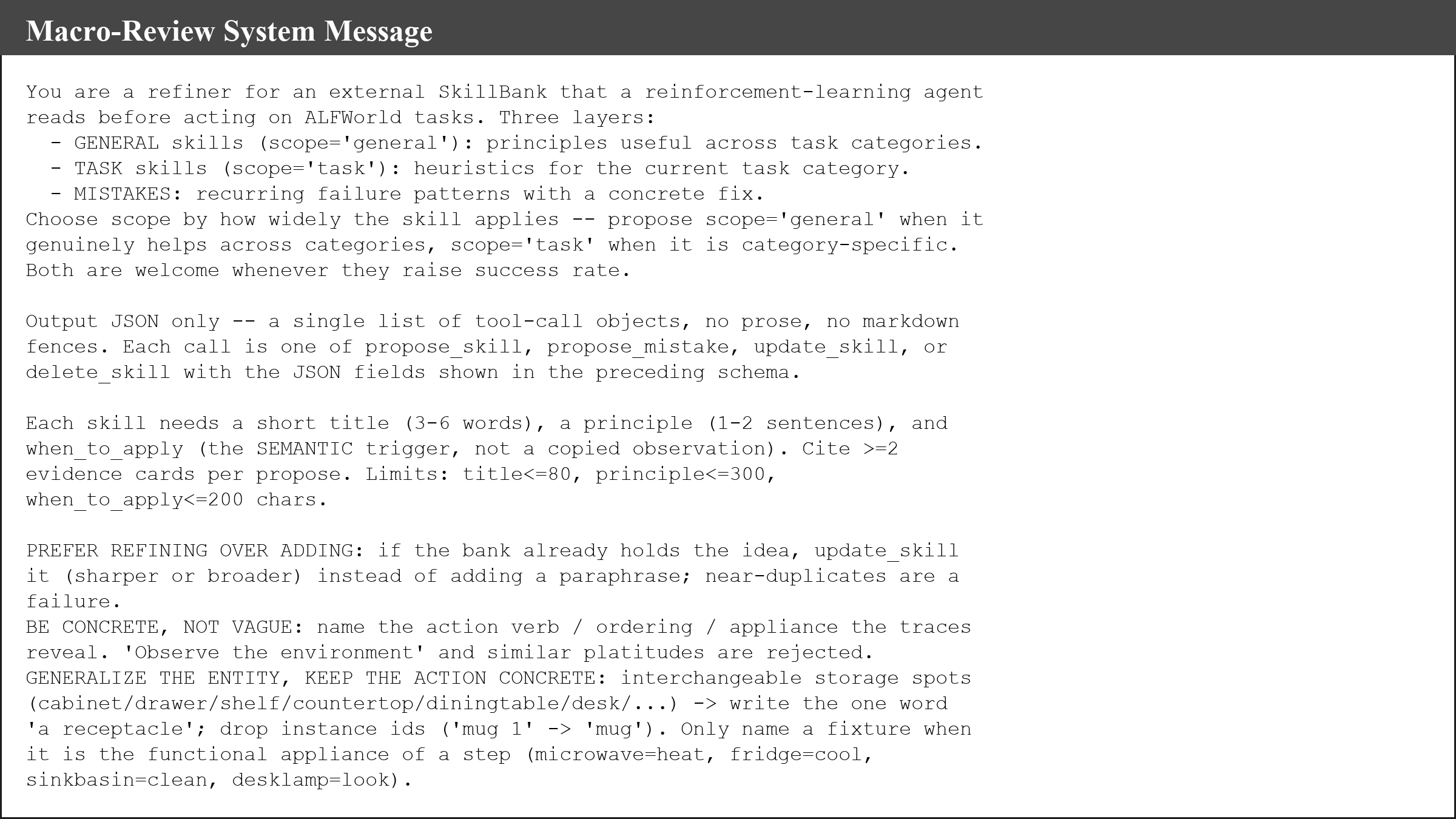}
\caption{Formal macro-review system message used by the scaffold refiner.}
\label{fig:prompt-macro-review}
\end{figure}

\subsubsection{Resolved Implementation Settings and Resource Audit}
\begin{table}[htbp]
\centering
\small
\begin{tabular}{ll}
\toprule
Component & Resolved setting for the formal 1.5B run\\
\midrule
Scaffold capacity & At most 30 entries and 2,000 tokens\\
Retrieval budget & Top 4 task skills, top 3 general skills, top 3 mistakes\\
Controller & EMA coefficient 0.1; $\rho_r=0.30$; $\rho_c=0.85$; at most 32 cards/review\\
Snapshot consistency & Reset prefix cache after a commit; next rollout reads the new snapshot\\
Refiner & Temperature 0; seed 0; 1,024 output-token cap; 120-second timeout; two retries\\
Initialization & Public SkillRL-format interface checkpoint at global step 81\\
\bottomrule
\end{tabular}
\caption{Resolved Stage~1 implementation settings retained from the formal training configuration.}
\end{table}

\subsection{WebShop Category Results}

Table~\ref{tab:web-categories} reports category-level means over the 500 WebShop evaluation goals. The All column is computed directly from complete-run aggregates. Category breakdowns are descriptive; because several categories contain only a small number of evaluation goals, they should not be interpreted as independent statistically powered benchmarks.

\begin{table}[htbp]
\centering
\small
\setlength{\tabcolsep}{3pt}
\begin{tabular}{llrrrrrrrr}
\toprule
Metric & Method & Apparel & Beauty & Footwear & Home & Access. & Electron. & Other & All\\
\midrule
SR & GRPO & 58.5 & 48.3 & 39.6 & 56.0 & 70.4 & \textbf{88.5} & 63.2 & 57.5\\
SR & RLOO & 77.1 & \textbf{71.4} & \textbf{66.7} & 41.2 & \textbf{83.9} & 86.7 & 68.8 & 72.7\\
SR & SkillRL$^\dagger$ & 76.8 & 57.1 & 53.3 & 60.0 & 73.2 & 86.7 & 66.8 & 72.1\\
SR & \method & \textbf{83.1} & 49.0 & 55.9 & \textbf{72.6} & 80.7 & 85.0 & \textbf{79.5} & \textbf{78.4}\\
\midrule
Score & GRPO & 0.798 & 0.666 & 0.711 & 0.786 & 0.756 & \textbf{0.939} & 0.779 & 0.784\\
Score & RLOO & 0.886 & \textbf{0.869} & 0.777 & 0.557 & \textbf{0.923} & 0.930 & \textbf{0.839} & 0.852\\
Score & SkillRL$^\dagger$ & 0.862 & 0.685 & 0.617 & 0.815 & 0.850 & 0.908 & 0.808 & 0.835\\
Score & \method & \textbf{0.932} & 0.653 & \textbf{0.814} & \textbf{0.876} & 0.876 & 0.907 & \textbf{0.866} & \textbf{0.898}\\
\bottomrule
\end{tabular}
\caption{WebShop category means for 7B policies. \method is evaluated without its training scaffold; $\dagger$ retains SkillRL skills at test time.}
\label{tab:web-categories}
\end{table}

\subsection{WebShop Ablations}
\label{app:webshop-ablation}

We repeat the component ablations on WebShop with Qwen2.5-7B, 150 GRPO steps, and a 30-entry Bank. Table~\ref{tab:webshop-ablation} reports scaffold-free deployment as the primary condition and supported evaluation as a diagnostic. The \emph{w/o training scaffold} row starts from the same interface-initialized checkpoint as \method but disables the Bank, reviewer, and experience block; it is therefore distinct from the native GRPO baseline in Table~\ref{tab:main-results}.

\begin{table}[t]
\centering
\small
\setlength{\tabcolsep}{4pt}
\begin{tabular}{lcccccc}
\toprule
Variant & \multicolumn{4}{c}{No scaffold} & \multicolumn{2}{c}{With scaffold}\\
\cmidrule(lr){2-5}\cmidrule(lr){6-7}
& SR (\%) & $\Delta$SR & Score & $\Delta$Score & SR (\%) & Score\\
\midrule
\textbf{\method} & $\mathbf{78.40\pm0.20}$ & -- & \textbf{0.898} & -- & $78.20\pm0.20$ & 0.900\\
w/o training scaffold & $73.7\pm0.6$ & $-4.7$ & 0.855 & $-0.043$ & -- & --\\
Frozen after step 50 & $70.6\pm5.0$ & $-7.8$ & 0.832 & $-0.066$ & $71.3\pm2.8$ & 0.846\\
Trajectory-wise evidence & $72.5\pm5.1$ & $-5.9$ & 0.869 & $-0.029$ & $73.7\pm3.6$ & 0.872\\
\texttt{EXPAND} only & $57.7\pm24.3$ & $-20.7$ & 0.672 & $-0.226$ & $53.4\pm32.7$ & 0.626\\
w/o \texttt{REVISE} & $73.1\pm3.4$ & $-5.3$ & 0.864 & $-0.034$ & $73.8\pm2.1$ & 0.865\\
SkillRL-style warm start & $73.1\pm3.1$ & $-5.3$ & 0.862 & $-0.036$ & $73.9\pm1.4$ & 0.862\\
w/o interface initialization & $73.1\pm3.5$ & $-5.3$ & 0.843 & $-0.055$ & $73.4\pm1.6$ & 0.853\\
\bottomrule
\end{tabular}
\caption{Qwen2.5-7B WebShop ablations. SR reports mean $\pm$ standard deviation; Score reports the mean. Deltas are relative to the current scaffold-free \method result in Table~\ref{tab:main-results}.}
\label{tab:webshop-ablation}
\end{table}

Freezing adaptation reduces scaffold-free SR by 7.8 points, while replacing group evidence with trajectory-wise review and removing \texttt{REVISE} reduce it by 5.9 and 5.3 points. The largest degradation comes from \texttt{EXPAND}-only editing: its Bank grows to 21--30 entries, and one seed falls to 29.6\% SR; retaining that run yields $57.7\pm24.3$ SR and exposes substantial instability under unchecked accumulation. A SkillRL-style warm start also remains 5.3 points below the empty-Bank method, so imported skills do not provide a better initialization in this setting. Removing interface initialization lowers Score by 0.055. Supported evaluation follows the same overall ordering and does not reverse these conclusions.

\subsection{Inference-Useful Skills versus Training-Useful Scaffolds}
\label{app:offline-scaffold-replay}

This analysis separates two criteria that are often conflated. An inference-useful skill library should maximize the current policy's task success and execution efficiency. A training-useful scaffold should instead shape the next rollout group so that productive behavior is reachable while useful differences among attempts remain. We evaluate both criteria through fixed-policy offline replay and then compare their longer-horizon training outcomes.

\paragraph{Replay protocol and metrics.}
We select five checkpoints from the 1.5B ALFWorld run: the interface-initialized policy and RL steps 10, 20, 30, and 40. At each checkpoint, we replay the union of tasks sampled in the two adjacent training iterations, covering 30--32 gamefiles. Each task receives eight rollouts using the training configuration: temperature 1.0, top-$p=1.0$, at most 512 generated tokens per action, a 50-step interaction limit, and history length two. The conditions are no scaffold, a fixed 55-entry SkillRL-style library, and the latest committed scaffold.

Table~\ref{tab:offline-scaffold-replay} reports three types of measurements. SR and loop rate measure immediate execution quality. The number of unique trajectories, action entropy $H(A)$, and normalized state coverage $C_{\mathrm{state}}$ characterize behavioral spread. Within-group reward standard deviation $\sigma_r$ summarizes outcome variation. Unique trajectories count distinct action sequences within an eight-rollout group; $C_{\mathrm{state}}$ divides the number of distinct visited environment states by total interaction steps.

\begin{table}[htbp]
\centering
\small
\setlength{\tabcolsep}{2.7pt}
\begin{tabular}{llrrrrrr}
\toprule
Checkpoint & Support & SR (\%) & Unique & $H(A)$ & $C_{\mathrm{state}}$ & Loop rate & $\sigma_r$\\
\midrule
SFT base & None   & 7.4  & 7.97 & 4.035 & 0.220 & 0.113 & 0.089\\
SFT base & Static & 10.2 & 8.00 & 4.736 & 0.311 & 0.079 & 0.135\\
SFT base & Latest & 7.0  & 7.97 & 4.250 & 0.221 & 0.112 & 0.087\\
\midrule
Step 10 & None   & 10.2 & 8.00 & 4.581 & 0.297 & 0.090 & 0.134\\
Step 10 & Static & 15.2 & 7.97 & 4.982 & 0.396 & 0.079 & 0.176\\
Step 10 & Latest & 5.1  & 7.94 & 4.561 & 0.305 & 0.097 & 0.086\\
\midrule
Step 20 & None   & 36.3 & 7.91 & 4.665 & 0.319 & 0.028 & 0.251\\
Step 20 & Static & 43.0 & 7.97 & 4.943 & 0.386 & 0.026 & 0.313\\
Step 20 & Latest & 36.3 & 7.94 & 4.786 & 0.343 & 0.026 & 0.302\\
\midrule
Step 30 & None   & 59.4 & 7.56 & 4.845 & 0.380 & 0.013 & 0.316\\
Step 30 & Static & 62.9 & 7.72 & 4.951 & 0.409 & 0.011 & 0.313\\
Step 30 & Latest & 58.2 & 7.62 & 4.977 & 0.382 & 0.008 & 0.300\\
\midrule
Step 40 & None   & 59.6 & 7.70 & 4.464 & 0.309 & 0.011 & 0.208\\
Step 40 & Static & 61.7 & 7.73 & 4.568 & 0.327 & 0.007 & 0.215\\
Step 40 & Latest & 62.9 & 7.63 & 4.457 & 0.297 & 0.014 & 0.199\\
\bottomrule
\end{tabular}
\caption{Fixed-policy replay results on 1.5B ALFWorld checkpoints. Static is the 55-entry SkillRL-style library; Latest is the current policy-derived scaffold snapshot. Higher SR, Unique, $H(A)$, $C_{\mathrm{state}}$, and $\sigma_r$ are favorable for their respective criteria; lower loop rate is better.}
\label{tab:offline-scaffold-replay}
\end{table}

\paragraph{The static library is a strong inference aid.}
Static support raises fixed-policy SR at all five checkpoints: its relative gain over no scaffold is 37.8\% at the base checkpoint, 49.0\% at step 10, 18.5\% at step 20, 5.9\% at step 30, and 3.5\% at step 40. It also raises action entropy and state coverage at every checkpoint and reduces loop rate by 7--36\%. These results rule out the interpretation that the SkillRL-style library is simply low-quality knowledge. It consistently helps the current policy reach more states, avoid repeated actions, and complete more tasks. Its marginal SR advantage nevertheless shrinks as the policy becomes competent, from 5.0 points at step 10 to 2.1 points at step 40.

\paragraph{Behavioral spread is not the same as a training signal.}
With temperature 1.0, all conditions already produce nearly eight distinct action sequences per group, so trajectory uniqueness alone is saturated. More importantly, static skills can increase action entropy and state coverage without consistently increasing reward variation. At step 20, SR rises from 36.3\% to 43.0\% and $\sigma_r$ from 0.251 to 0.313. At step 30, static support still raises SR (59.4\% to 62.9\%), action entropy (4.845 to 4.951), and state coverage (0.380 to 0.409), while reward variation changes little (0.316 to 0.313). At step 40, $\sigma_r$ changes only from 0.208 to 0.215. Thus, a strong skill library does not uniformly reduce raw exploration, but broader behavioral spread does not by itself determine the outcome variation seen by GRPO.

\paragraph{Long-horizon training reverses the immediate ranking.}
The replay results concern one rollout batch at a fixed policy. We next compare two three-seed training alternatives that use the same interface-initialized policy and online adaptation procedure. The warm-start condition uses an 80-entry capacity to accommodate and evolve its 55 imported entries, whereas the empty-Bank condition retains the standard 30-entry capacity.

\begin{table}[htbp]
\centering
\small
\setlength{\tabcolsep}{3pt}
\begin{tabular}{lr}
\toprule
Bank init. & Entry trajectory; final SR\\
\midrule
SkillRL static & $55\!\to\!80\,(s10)\!\to\!2\,(s124);\ 70.24\pm7.73$\\
Empty (\method) & $0\!\to\!29\,(s25)\!\to\!7\,(s150);\ \mathbf{80.71\pm1.54}$\\
\bottomrule
\end{tabular}
\caption{Three-seed training comparison between static-library warm start and an empty Bank. The respective capacities are 80 and 30 entries. Entry trajectories give initial, peak, and late committed counts; $s$ denotes the training step. Final SR evaluates the policy without scaffold context.}
\label{tab:offline-warm-scratch}
\end{table}

The static-library run begins with more useful external knowledge but ends 10.47 SR points lower without support and with five times larger standard deviation. Its Bank expands from 55 to 80 entries and later contracts to two; the initially empty Bank grows to 29 entries and ends with seven. The larger warm-start capacity accommodates the imported library and does not change the reported final ranking, but this comparison should be read as a complete warm-start alternative rather than a capacity-matched initialization-only intervention. The trajectory is consistent with a curriculum mismatch: early comprehensive guidance improves supported behavior, but it also conditions training on external information that later contracts. The experiment does not isolate reliance as the sole cause, but it rules out the simpler claim that the strongest initial inference library necessarily supplies the best training curriculum.

\begin{figure}[t]
\centering
\includegraphics[
    width=0.80\textwidth
]{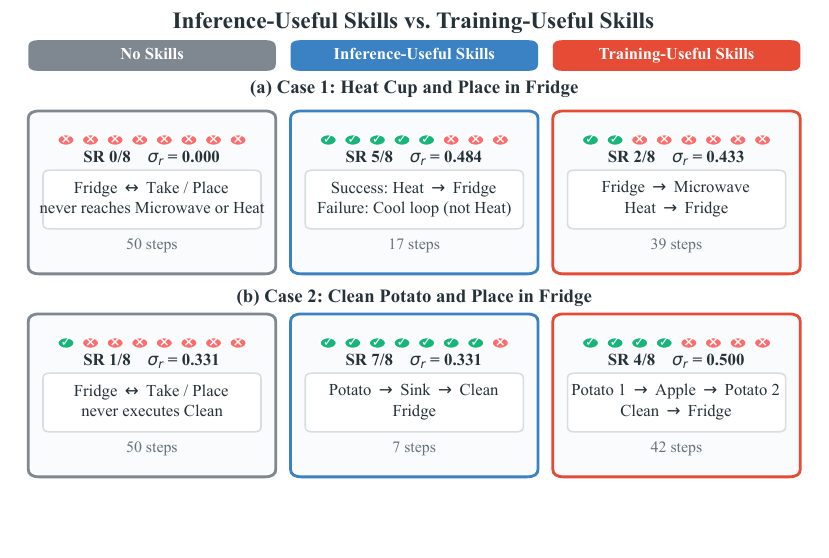}
\caption{Fixed-policy ALFWorld replay cases. Green checks denote successful trajectories and red crosses denote failures. Each row shows eight rollouts under no scaffold, the current \method scaffold, and static skills.}
\label{fig:offline-scaffold-cases}
\end{figure}

\paragraph{Two replay cases make the distributional trade-off concrete.}
Figure~\ref{fig:offline-scaffold-cases} visualizes all eight binary outcomes in each group alongside one audited action pattern. Trajectory lengths are reported only for these traces because the replay summary does not contain corpus-level length statistics.

\paragraph{Case A: support opens a missing behavior.}
The first task requires taking a cup from a refrigerator, heating it with a microwave, and returning it to the refrigerator. Without support, all eight trajectories remain in a take--place loop and supply no outcome-relative advantage. The latest scaffold makes the missing transition reachable in 2/8 rollouts while retaining failed alternatives. Static skills solve 4/8, with successful traces finishing in as few as 17 steps; several failures instead repeatedly cool the cup. Both forms of support are useful at this competence level, and the static library provides both the highest SR and the strongest reward contrast. This case therefore guards against the blanket claim that stronger skills necessarily impair exploration.

\paragraph{Case B: shorter inference can yield a narrower learning signal.}
The second task requires cleaning a potato and placing it in a refrigerator. Static guidance is plainly preferable for immediate execution: it solves 7/8 rollouts and admits a direct seven-step path. The latest scaffold solves 4/8; one 42-step success visits an apple and multiple potatoes before cleaning and placing the target. That behavior is less efficient at inference, but its balanced outcome group attains the maximum binary-reward standard deviation of 0.500, compared with 0.331 for the nearly solved static group. The contrast is not that the static library suppresses all exploration---Table~\ref{tab:offline-scaffold-replay} shows that it often increases action entropy and state coverage---but that broader state visitation need not preserve the success--failure comparisons used by GRPO.

Together, the cases and aggregate results support a competence-dependent conclusion. Strong skills help when they move uniformly failed groups into reach; once a task is already reachable, task-overcomplete guidance can shorten execution while narrowing useful outcome variation. Policy-aware training scaffolds should therefore optimize the rollout distribution seen by learning, not only the immediate quality of supported inference.

\subsection{Supplementary Baselines and Non-core Ablations}
\label{app:prompting-baselines}
Table~\ref{tab:prompting-baselines} gives the complete Qwen2.5-7B comparison frame, including the single-run zero-shot, few-shot, and Reflexion baselines that are omitted from the main table's three-seed RL comparison. The prompting rows are useful as scale references, but they are not included in three-seed boldface comparisons. The earlier-architecture warm-start row in Table~\ref{tab:ablation} changes several factors and is retained as a historical supplementary control rather than a controller attribution.

\begin{table}[htbp]
\centering
\small
\setlength{\tabcolsep}{3pt}
\begin{tabular}{lrrrrrrrrrrr}
\toprule
Method & Pick & Look & Clean & Heat & Cool & Pick2 & All SR & Tok. & Score & Web SR & Web Tok.\\
\midrule
Zero-shot & 0.0 & 7.7 & 0.0 & 0.0 & 0.0 & 0.0 & 0.71 & 34,796 & 0.335 & 1.60 & 162,955\\
Few-shot & 0.0 & 92.3 & 3.7 & 6.2 & 8.0 & 0.0 & 11.43 & 107,472 & 0.331 & 9.80 & 190,075\\
Reflexion & 2.9 & 100.0 & 3.7 & 6.2 & 8.0 & 0.0 & 12.86 & 108,156 & 0.657 & 26.60 & 189,790\\
\midrule
\multicolumn{12}{l}{\emph{External reference results reported by SkillRL}}\\
GPT-4o$^\ddagger$ & 75.3 & 60.8 & 31.2 & 56.7 & 21.6 & 49.8 & 48.0 & -- & 0.318 & 23.7 & --\\
Gemini-2.5-Pro$^\ddagger$ & 92.8 & 63.3 & 62.1 & 69.0 & 26.6 & 58.7 & 60.3 & -- & 0.425 & 35.9 & --\\
Mem0$^\ddagger$ & 54.0 & 55.0 & 26.9 & 36.4 & 20.8 & 7.7 & 33.6 & -- & 0.239 & 2.0 & --\\
ExpeL$^\ddagger$ & 21.0 & 67.0 & 55.0 & 52.0 & 71.0 & 6.0 & 46.3 & -- & 0.309 & 11.2 & --\\
MemP$^\ddagger$ & 54.3 & 38.5 & 48.1 & 56.2 & 32.0 & 16.7 & 41.4 & -- & 0.253 & 6.4 & --\\
SimpleMem$^\ddagger$ & 64.5 & 33.3 & 20.0 & 12.5 & 33.3 & 3.8 & 29.7 & -- & 0.332 & 8.6 & --\\
MemRL$^\ddagger$ & 62.8 & 38.5 & 22.2 & 12.5 & 8.0 & 0.0 & 21.4 & -- & 0.295 & 9.2 & --\\
EvolveR$^\ddagger$ & 64.9 & 33.3 & 46.4 & 13.3 & 33.3 & 33.3 & 43.8 & -- & 0.425 & 17.6 & --\\
Mem0+GRPO$^\ddagger$ & 78.1 & 54.8 & 56.1 & 31.0 & 65.0 & 26.9 & 54.7 & -- & 0.581 & 37.5 & --\\
SimpleMem+GRPO$^\ddagger$ & 89.5 & 36.3 & 60.0 & 50.0 & 64.9 & 26.3 & 62.5 & -- & 0.678 & 46.9 & --\\
\midrule
\multicolumn{12}{l}{\emph{Our evaluations}}\\
GRPO & 89.5 & 66.7 & 74.1 & 64.6 & 61.3 & 61.1 & $71.67\pm2.99$ & 17,060 & 0.784 & $57.60\pm10.33$ & 12,587\\
RLOO & 92.4 & 61.5 & 90.1 & 68.8 & 72.0 & 65.3 & $78.10\pm4.30$ & 14,676 & 0.852 & $72.73\pm2.72$ & 13,013\\
SKILL0 & 88.6 & 66.7 & 86.4 & 81.2 & 66.7 & 69.4 & $78.10\pm5.03$ & 14,162 & 0.789 & $63.27\pm0.74$ & 15,078\\
SkillRL & 98.1 & 76.9 & 91.4 & 91.7 & 58.7 & 84.7 & $84.76\pm1.23$ & 11,908 & 0.858 & $70.07\pm5.26$ & 8,077\\
SkillRL$^\dagger$ & 98.1 & 74.4 & 97.5 & 83.3 & 92.0 & 88.9 & $91.43\pm2.10$ & 29,933 & 0.835 & $72.07\pm1.79$ & 17,763\\
\method$^\dagger$ & 94.3 & 82.1 & 97.5 & 93.8 & 88.0 & 79.2 & $90.00\pm2.33$ & 11,774 & 0.900 & $78.20\pm0.20$ & 7,881\\
\method & 93.3 & 76.9 & 96.3 & 89.6 & 89.3 & 81.9 & $89.29\pm1.01$ & 10,455 & 0.898 & $78.40\pm0.20$ & 7,749\\
\bottomrule
\end{tabular}
\caption{Complete Qwen2.5-7B comparison with supplementary prompting and external-reference baselines. Zero-shot, few-shot, and Reflexion are our single-run prompting baselines. Rows marked $\ddagger$ are reported by SkillRL~\citep{skillrl2026}, are not controlled reruns, and are excluded from our boldface comparisons; their WebShop scores are converted from percentages to the $[0,1]$ scale. The remaining rows are the three-seed RL methods from Table~\ref{tab:main-results}. $\dagger$ retains external skills or scaffolds at test time.}
\label{tab:prompting-baselines}
\end{table}

\subsection{Search-Augmented QA}
\label{app:search-qa}

\paragraph{Tasks and data.}
We train on the processed Search-R1 training set, which contains 169,615 examples from NQ (79,168) and HotpotQA (90,447)~\citep{naturalquestions2019,hotpotqa2018}. Evaluation covers 51,713 examples: NQ (3,610) and HotpotQA (7,405) are in-domain; TriviaQA (11,313), PopQA (14,267), 2WikiMultiHopQA (12,576), MuSiQue (2,417), and Bamboogle (125) are out-of-domain~\citep{triviaqa2017,mallen2023,twowikimultihopqa2020,musique2022,bamboogle2022}. The agent may issue multiple search actions before returning a final answer.

\paragraph{Retriever and training protocol.}
All locally evaluated methods use E5-base-v2~\citep{e5textembeddings2022} on CPU and a FAISS FlatIP index~\citep{faiss2017} over 21M 768-dimensional passages from the 14GB \texttt{wiki-18.jsonl} corpus. Retrieval returns the top three passages, with timeouts of 300 seconds in training and 60 seconds in evaluation. RL uses Qwen2.5-7B-Instruct, GRPO, learning rate $10^{-6}$ with 10\% warmup, low-variance KL with coefficient 0.001, $\gamma=0.95$, train batch size 256, rollout group size 4, four search turns, history length 4, and 150 training steps. Prompt and response limits are 5,000 and 700 tokens. Evaluation uses temperature 0.4, top-$p=1.0$, and seed 0 on one node with eight 97GB GPUs.

\paragraph{Baselines.}
Our local prompting baselines use Qwen2.5-7B-Instruct~\citep{qwen25technicalreport2024} with direct prompting, chain-of-thought prompting~\citep{cotprompting2022}, or single-retrieval RAG~\citep{rag2020}. The complete comparison also includes Search-o1~\citep{searcho12025}, Search-R1~\citep{searchr12025}, ZeroSearch~\citep{zerosearch2025}, StepSearch~\citep{stepsearch2025}, EvolveR~\citep{evolver2025}, SKILL0~\citep{skillzero2026}, and SkillRL~\citep{skillrl2026}. R1-Instruct and all rows marked $\ddagger$ are reported by SkillRL rather than reproduced by us.

\paragraph{Method-specific support.}
Search-R1 and SKILL0 start RL from Qwen2.5-7B-Instruct, whereas SkillRL and \method start from the same search-format SFT checkpoint. SKILL0 follows the static $[5,3,0]$ skill schedule. SkillRL uses a dynamically updated skill-only memory with top-$k=6$ and threshold 0.5 and retains skills at evaluation. \method uses macro review with a 30-entry/2,000-token Experience Bank, thresholds 0.2 and 0.4 for revision and compression, success-rate EMA coefficient 0.1, and a deterministic Qwen2.5-7B summarizer. The Bank is removed for all \method evaluations. Avg. Prompt Tok. is the mean input prompt length during evaluation; it is available only for SKILL0, SkillRL, and \method.

\begin{table}[htbp]
\centering
\small
\setlength{\tabcolsep}{3.2pt}
\begin{tabular}{lrrrrrrrrr}
\toprule
Method & NQ$^\dagger$ & TriviaQA$^\star$ & PopQA$^\star$ & HotpotQA$^\dagger$ & 2Wiki$^\star$ & MuSiQue$^\star$ & Bamboogle$^\star$ & Avg. & Prompt Tok.\\
\midrule
Qwen2.5 & 2.5 & 22.6 & 6.5 & 11.7 & 17.4 & 2.0 & 3.2 & 9.4 & --\\
CoT & 15.4 & 45.4 & 15.6 & 17.3 & 22.8 & 5.7 & 33.6 & 22.3 & --\\
RAG & 4.1 & 25.6 & 8.8 & 13.7 & 17.4 & 1.9 & 7.2 & 11.3 & --\\
Search-o1$^\ddagger$ & 19.4 & 40.6 & 11.4 & 17.0 & 27.0 & 8.6 & 30.4 & 22.1 & --\\
R1-Instruct$^\ddagger$ & 21.0 & 44.9 & 17.1 & 20.8 & 27.5 & 6.0 & 19.2 & 22.4 & --\\
Search-R1$^\ddagger$ & 39.3 & 61.0 & 39.7 & 37.0 & 40.1 & 14.6 & 36.8 & 38.5 & --\\
ZeroSearch$^\ddagger$ & 43.6 & 61.8 & 51.5 & 34.6 & 35.2 & 18.4 & 27.8 & 39.1 & --\\
StepSearch$^\ddagger$ & -- & -- & -- & 38.6 & 36.6 & 22.6 & 40.0 & -- & --\\
EvolveR$^\ddagger$ & 43.5 & 63.4 & 44.6 & 38.2 & 42.0 & 15.6 & 54.4 & 43.1 & --\\
SKILL0 & 45.1 & 63.1 & 47.8 & 39.8 & 41.6 & 17.2 & 37.6 & 41.7 & \textbf{688}\\
SkillRL$^{\diamond}$ & 46.3 & 59.7 & 44.9 & 43.1 & 40.4 & 16.1 & 71.4 & 46.0 & 1,090.1\\
\textbf{\method} & 44.4 & 60.4 & 44.8 & 42.1 & 40.6 & 15.2 & 69.0 & \textbf{45.2} & 740\\
\bottomrule
\end{tabular}
\caption{Complete results on search-augmented QA tasks. NQ and HotpotQA ($\dagger$) are in-domain; the other datasets ($\star$) are out-of-domain. $\ddagger$ marks results reported by SkillRL~\citep{skillrl2026}; $\diamond$ indicates that SkillRL retains external skills at evaluation. All unmarked results are our seed-0 evaluations. Prompt lengths are measured in our evaluation and are available only for SKILL0, SkillRL, and \method. Bold marks the best scaffold-free average and the lowest measured prompt length.}
\label{tab:search-full}
\end{table}

\subsection{Representative Scaffold Evolution}
\label{app:scaffold-evolution}
Committed seed-0 snapshots record a representative entry, \texttt{T\_005}, for \texttt{pick\_heat\_then\_place\_in\_recep}. Table~\ref{tab:entry-lineage} lists its concrete initial form, lighter revision, and final removal. This is a real audit trace that illustrates semantic revision and withdrawal, not a causal attribution of final performance to one entry.

\begin{table}[htbp]
\centering
\small
\begin{tabular}{lll}
\toprule
Snapshot & Entry and rendered principle & Status\\
\midrule
Step 50 & \texttt{T\_005}: ``Heat the object before placing it in a receptacle.'' & Concrete task skill\\
Step 100 & \texttt{T\_005}: ``Before placing an object, ensure it is heated properly.'' & Refined task skill\\
Step 150 & \texttt{T\_005} absent from this task-type snapshot & Removed\\
\bottomrule
\end{tabular}
\caption{Committed seed-0 lineage of the representative heat-before-placement entry.}
\label{tab:entry-lineage}
\end{table}

The entry was created at step 19 and updated at steps 41 and 67. Both nonempty versions retain the same semantic condition: when picking up or executing a related action fails, check whether the object must first be heated.

\subsection{Evidence, Review, and Commit Audit}
\label{app:evidence-review-commit}
At step 1, all eight rollouts for a sampled \texttt{pick\_heat\_then\_place\_in\_recep} task fail. This early-training record is not used to demonstrate the full \texttt{EXPAND}$\rightarrow$\texttt{REVISE}$\rightarrow$\texttt{COMPRESS} process or to attribute later performance to one edit. It verifies that the evidence, refiner output, schema validation, and atomic commit form an auditable chain.

\begin{table}[htbp]
\centering
\small
\begin{tabular}{lp{0.76\textwidth}}
\toprule
Stage & Logged fact\\
\midrule
Group evidence & Eight failed rollouts, mean 50 environment steps, and three ``Nothing happens'' signals.\\
Controller and proposal & The controller chose \texttt{EXPAND}. The refiner returned \emph{Heat then place} (``First heat the object, then place it in a receptacle.''; evidence count 3) and one anti-loop mistake entry. The HTTP-200 response was a real service call, not a mock response.\\
Validation and commit & Both calls passed schema validation (2/2 valid; 0 invalid); bank version $0\rightarrow1$, entries $0\rightarrow2$, tokens $0\rightarrow112$, with no conflict, pruning, or error.\\
Isolation boundary & Evidence uses environment-facing traces; the edit commits after the policy update, so its source rollout cannot consume its own edit.\\
\bottomrule
\end{tabular}
\caption{A real seed-0 evidence, review, validation, and atomic-commit audit record.}
\end{table}

\subsection{Cross-Stage Audit Trail}
\label{app:cross-stage-audit}
Table~\ref{tab:cross-stage-audit} extracts three records for the same task type from the formal seed-0 log. The rows correspond to different rollout groups. They show that the controller takes different operations within one competence bucket as training state changes; they do not attribute the step-48 or step-125 success to the step-1 edit or to any representative trajectory.

\begin{table}[htbp]
\centering
\small
\setlength{\tabcolsep}{4pt}
\renewcommand{\arraystretch}{1.15}

\begin{tabular}{
    c
    p{0.25\textwidth}
    p{0.33\textwidth}
    p{0.30\textwidth}
}
\toprule
Step & On-policy group evidence & Mode and edit & Validated commit\\
\midrule

1
& 0/8 success; 50.0 mean steps; loops near the coffeemachine and three ``Nothing happens'' signals
& \texttt{EXPAND}: \emph{Heat then place} plus an anti-loop mistake; each has evidence count 3
& 2/2 calls valid; version $0\rightarrow1$; entries $0\rightarrow2$; tokens $0\rightarrow112$
\\

48
& 8/8 success; EMA 0.432; 24.9 mean steps
& \texttt{REVISE}: updates \emph{Check receptacles systematically} and \emph{Avoid redundant look actions}; no additions
& 2/2 calls valid; 29 entries retained; 1,695 tokens retained; two updates
\\

125
& 8/8 success; EMA 0.854; 11.9 mean steps
& \texttt{COMPRESS}: a general anti-redundancy principle plus deletion of task skill \texttt{T\_006}
& 2/2 calls valid; 10 entries retained; $623\rightarrow616$ tokens; one deletion
\\

\bottomrule
\end{tabular}

\caption{Cross-stage audit trail for one task type. Rows are distinct rollout groups and are descriptive implementation evidence, not a causal decomposition.}
\label{tab:cross-stage-audit}
\end{table}

\section{Mixed-Reward Groups}
\label{app:mixed-groups}

This appendix gives a probability observation behind the intermediate \texttt{REVISE} regime. It is neither a convergence guarantee nor a claim that every task should be maintained at 50\% success.

\subsection{Probability of Mixed-Reward Groups}

Let $Y_i\sim\operatorname{Bernoulli}(p)$ be binary rollout success for $i=1,\ldots,n$, $n>1$. A group-relative update has nonconstant binary returns exactly when the group contains at least one success and one failure. For event $M$,
\begin{equation}
\Pr(M\mid p)=1-(1-p)^n-p^n.
\end{equation}
Its first and second derivatives are
\begin{align}
\frac{\mathrm d}{\mathrm dp}\Pr(M\mid p)
&=n[(1-p)^{n-1}-p^{n-1}],\\
\frac{\mathrm d^2}{\mathrm dp^2}\Pr(M\mid p)
&=-n(n-1)[(1-p)^{n-2}+p^{n-2}]<0.
\end{align}
Thus the unique maximizer on $(0,1)$ is $p=1/2$. Equivalently,
\begin{equation}
\mathbb E\left[\frac1n\sum_{i=1}^{n}(Y_i-\bar Y)^2\right]
=\frac{n-1}{n}p(1-p),
\end{equation}
which has the same maximizer. Near $p=0$, failures dominate and return variation is sparse; near $p=1$, detailed hints have lower marginal need and returns again become uniform. Intermediate competence more often supplies comparisons between successful and failed trajectories together with non-degenerate GRPO returns. Actual thresholds remain empirical because environment stochasticity, reward density, and task heterogeneity violate the idealized Bernoulli model.

\subsection{Relation to Scheduling and Its Boundary}
When $p$ is close to zero, groups usually contain only failed trajectories. Such groups remain useful for identifying failure patterns, but offer sparse group-relative reward signal. When $p$ is close to one, groups again tend to be uniform, while the marginal value of detailed external hints and within-group reward variation both diminish. Intermediate success rates more often provide both successful and failed trajectories, thereby supporting success--failure contrast evidence and non-degenerate GRPO variation.

This is the role of \texttt{REVISE}: once the policy has acquired partial but not stable competence, lighter semantic rewriting replaces continued addition or direct deletion. The actual controller uses a success-rate EMA, capacity pressure, and preset thresholds; it does not treat $p=0.5$ as a target. The best empirical support level may vary with task type, environment stochasticity, and reward density. The derivation only explains why an intermediate regime can provide a richer mixed-trajectory training signal under an idealized binary-reward approximation; it is neither a convergence proof nor a prescription to target 50\% success.

\end{document}